\documentclass{article}

\usepackage{microtype}
\usepackage{graphicx}
\usepackage{subcaption}
\usepackage{booktabs} 

\usepackage{hyperref}

\usepackage[preprint]{icml2026}

\usepackage{amsmath, amssymb, mathtools}
\usepackage{amsthm}
\usepackage{xcolor}
\usepackage{graphicx}
\usepackage{caption} 
\usepackage{subcaption}
\usepackage{multirow}
\usepackage{adjustbox}
\usepackage{xspace}
\usepackage{enumitem}
\usepackage{colortbl}
\usepackage{listings}

\usepackage{url}
\usepackage{booktabs}
\usepackage[most]{tcolorbox}
\usepackage{hyperref}
\usepackage[capitalize,noabbrev]{cleveref}

\usepackage{tikz}
\usetikzlibrary{arrows.meta,positioning,decorations.pathreplacing,calc}

\definecolor{melon}{HTML}{F89E7B}

\newcommand{\method}{\textsc{CompAs}\xspace}
\newcommand{\wilda}{\textsc{wilda}\xspace}

\newcommand{\nospacetext}[1]{\makebox[0pt][l]{#1}}

\definecolor{myblue}{RGB}{21,101,192}
\definecolor{myred}{RGB}{198,40,40}

\theoremstyle{plain}
\newtheorem{theorem}{Theorem}[section]

\newtheorem{corollary}[theorem]{Corollary}
\theoremstyle{definition}

\theoremstyle{remark}

\usepackage[textsize=tiny]{todonotes}

\icmltitlerunning{Context Parametrization with Compositional Adapters}

\begin{document}

\twocolumn[
  \icmltitle{Context Parametrization with Compositional Adapters}



  \icmlsetsymbol{equal}{*}

  \begin{icmlauthorlist}
    \icmlauthor{Josip Juki{\'{c}}}{fer}
    \icmlauthor{Martin Tutek}{fer}
    \icmlauthor{Jan {\v{S}}najder}{fer}
  \end{icmlauthorlist}

 \icmlaffiliation{fer}{TakeLab, University of Zagreb, Croatia}

 \icmlcorrespondingauthor{Josip Juki{\'{c}}}{josip.jukic@fer.hr}

  \icmlkeywords{LLMs}

  \vskip 0.3in
]



\printAffiliationsAndNotice{}  

\begin{abstract}
Large language models (LLMs) often seamlessly adapt to new tasks through in-context learning (ICL) or supervised fine-tuning (SFT). However, ICL is inefficient when handling many demonstrations, and SFT incurs training overhead while sacrificing flexibility.  
Mapping instructions or demonstrations from context directly into adapter parameters offers an appealing alternative. While prior work explored generating adapters based on a single input context, it has overlooked the need to integrate multiple chunks of information.
To address this gap, we introduce \method, a meta-learning framework that translates context into adapter parameters with a compositional structure that allows them to be merged algebraically.  
This approach yields three benefits: lower inference cost, improved stability under long contexts, and establishes a principled solution when input exceeds the model’s context window.  
Furthermore, \method reversibly encodes information into adapter parameters, enabling recovery of the original input context and facilitating safety.
Empirical results on diverse multiple-choice and extractive question answering tasks show that \method outperforms ICL and prior generator-based methods, especially when scaling to more inputs. 
Our work establishes composable adapter generation as a practical and efficient alternative for scaling LLM deployment.
\end{abstract}


\section{Introduction}

Large language models (LLMs) adapt to new tasks by integrating \emph{contextual information} pertaining to these tasks, such as instructions, demonstrations, or extracted evidence.  
This adaptability is typically realized through in-context learning (ICL) \citep{brown2020language}, instruction following \citep{ouyang2022instructgpt}, and retrieval-augmented generation \citep{lewis2020retrieval, borgeaud2022improving}, where extra tokens in the prompt act as a transient memory that steers model behavior.  
Another common strategy is to adapt parameters directly through \emph{supervised fine-tuning} (SFT), ranging from full model updates \citep{devlin-etal-2019-bert, raffel2020exploring} to parameter-efficient variants \cite{pfeiffer2023modular}.  
While effective, both prompting and SFT face limitations.  
Prompt-based methods often require long contexts, which destabilize attention and degrade performance as length increases \citep{liu-etal-2024-lost}, while also incurring inference costs that scale with context size.  
SFT, on the other hand, requires additional training overhead and lacks flexibility, as an individual model is trained for each task.

A complementary line of work views \emph{adapters} \citep{houlsby2019parameter, hu2022lora} as a mechanism for parametrizing context. In our work, we use \textit{context} as an umbrella term for instructions, demonstrations, or supporting passages.
By translating context into parameters, adapters offer a persistent representation that replaces prompt tokens, thereby improving the stability of processing long contexts and amortizing adaptation across tasks \citep{karimi2021compacter, he2022towards, liu-etal-2024-lost}.  
Once context is encoded into adapters, inference requires only the query tokens, reducing the cost of processing long prompts and making the approach especially attractive in latency- or memory-constrained settings.  
This context-to-adapter transformation encodes an otherwise transient token sequence into a manipulable object, allowing for caching, reuse, and algebraic combination.

Building on this perspective, recent work has leveraged meta-learning to efficiently generate adapter parameters directly from support examples without fine-tuning anew for each context \citep{bansal2022metaadapters, chen2025generativeadapter}. 
\begin{figure*}[]
    \centering
    \begin{tikzpicture}
[
  font=\small, node distance=4mm and 8mm,
  teacherbox/.style={rectangle, rounded corners, draw, minimum height=9mm, minimum width=42mm, fill=gray!30, align=center, line width=0.6mm},
  studentbox/.style={rectangle, rounded corners, draw, minimum height=19mm, minimum width=62mm, fill=gray!10, align=center, line width=0.6mm},
  genbox/.style={rectangle, rounded corners, draw, minimum height=13mm, minimum width=40mm, fill=gray!20, align=center, line width=0.6mm, inner sep=1.6mm},
  basebox/.style={rectangle, rounded corners, draw, minimum height=6.2mm, minimum width=18mm, align=center, fill=gray!5, line width=0.4mm},
  adapterS/.style={rectangle, rounded corners, draw=orange!80!black, densely dotted, minimum height=6.2mm, minimum width=12mm, align=center, fill=orange!24, line width=0.6mm},
  adapterG/.style={rectangle, rounded corners, draw, minimum height=6.2mm, minimum width=16mm, align=center, fill=gray!5, line width=0.4mm},
  demobox/.style={
    rectangle, rounded corners, draw=myblue!70,
    minimum height=7mm, minimum width=20mm,
    align=center, fill=myblue!15, line width=0.6mm
  },
  querybox/.style={
    rectangle, rounded corners, draw=myred!70,
    minimum height=7mm, minimum width=20mm,
    align=center, fill=myred!15, line width=0.6mm
  },
  arrCtx/.style={-{Triangle[length=1.6mm,width=1.3mm]}, line width=0.8pt, draw=black!80, shorten >=1pt, shorten <=1pt},
  arrGdot/.style={-{Triangle[length=1.6mm,width=1.3mm]}, line width=1.8pt, dotted, draw=green!60!black},
  arrGen/.style={-{Triangle[length=1.6mm,width=1.3mm]}, line width=1pt, dotted, draw=orange!60!black}
]

\node[demobox] (c1) {Context $\mathbf{c}_1$};
\node[demobox, below=2mm of c1] (c2) {Context $\mathbf{c}_2$};
\node at ([yshift=-6mm]c2) (cdots) {\Large $\vdots$};
\node[demobox, below=6mm of c2] (ck) {Context $\mathbf{c}_k$};

\node[genbox, anchor=south west] (G) at ($(ck.south east)+(4mm,0mm)$) {};
\node[anchor=north west] at ([xshift=1.2mm,yshift=-0.8mm]G.north west) {\textbf{Generator $G$}};
\node[basebox, anchor=east]   (Gbase) at ($(G.center)+( 0mm,-2.2mm)$) {Base LLM};
\node[adapterG, anchor=west]  (Gadp)  at ($(G.center)+(  2mm,-2.2mm)$) {Adapter};

\node[querybox, anchor=south] (q) at ($(G.north)+(0,2mm)$) {Query $\mathbf{q}$};

\node[studentbox, anchor=south west] (S) at ($(ck.south east)+(48mm,0mm)$) {};
\node[anchor=north west] at ([xshift=1.2mm,yshift=-0.8mm]S.north west) {\textbf{Student $S$}};
\node[basebox] (Sbase) at ($(S.north)+(0,-5mm)$) {Base LLM};

\node[adapterS, below=3mm of Sbase, xshift=-22.5mm] (Ha1) {Adapter$_1$};
\node[circle, draw=black!70, fill=white, inner sep=0.1mm, minimum size=2.6mm, right=0.5mm of Ha1] (hplus1) {\scriptsize $+$};
\node[adapterS, right=0.5mm of hplus1] (Ha2) {Adapter$_2$};
\node[right=0.5mm of Ha2] (hdots) {\footnotesize $\cdots$};
\node[circle, draw=black!70, fill=white, inner sep=0.1mm, minimum size=2.6mm, right=0.5mm of hdots] (hplusk) {\scriptsize $+$};
\node[adapterS, right=0.5mm of hplusk] (Hak) {Adapter$_k$};

\node[teacherbox, anchor=south west] (T) at ($(S.north west)+(0,1mm)$) {};
\node[anchor=north west] at ([xshift=1.2mm,yshift=-0.6mm]T.north west) {\textbf{Teacher $T$}};
\node[basebox, anchor=center] (Tbase) at ($(T.center)+(8mm,0)$) {Base LLM};

\node[circle, draw=black!70, inner sep=0.5mm, minimum size=4mm,
      anchor=south] (concat) at ($(q.north)+(0,3mm)$) {$;$};

\coordinate (GinTop)    at ($(G.west)+(0, 4mm)$);
\coordinate (GinMiddle) at ($(G.west)+(0, 0mm)$);
\coordinate (GinBottom) at ($(G.west)+(0,-3mm)$);

\draw[arrCtx] ($(c1.east) + (0, -1mm)$) to[out=-60,in=120] (GinTop);
\draw[arrCtx] ($(c2.east) + (0, -2mm)$) to[out=-60,in=160] (GinMiddle);
\draw[arrCtx] (ck.east) -- (GinBottom);

\draw[arrGen] ($(G.east)+(0,-3mm)$) to[out=0,in=180] ($(Ha1.west)+(0,-1mm)$);
\draw[arrGen] ($(G.east)+(0,3mm)$) to[out=0,in=170] ($(Ha2.north)+(-6.5mm,0)$);
\draw[arrGen] ($(G.east)+(0,5mm)$) to[out=0,in=175] ($(Hak.north)+(-6.5mm,0)$);

\draw[-{Triangle[length=1.6mm,width=1.3mm]}, semithick, draw=black!80]
  (q.east) to[out=0,in=180] ($(S.west)+(0,5mm)$);

\draw[-{Triangle[length=1.6mm,width=1.3mm]}, line width=0.8pt, draw=gray!80]
  (c1.east) |- (concat.west);

\draw[-{Triangle[length=1.6mm,width=1.3mm]}, line width=0.8pt, draw=gray!80]
  (q.north) -- (concat.south);

\draw[-{Triangle[length=1.6mm,width=1.3mm]}, line width=0.8pt, draw=gray!80]
  (concat.east)  to[out=0,in=180] (T.west) ;

\node[right=10mm of T.east] (yt) {$\mathbf{y}_t$};
\node[right=3mm of S.east] (ys) {$\mathbf{y}_s$};

\draw[-{Triangle[length=1.8mm,width=1.5mm]}, thick, draw=black!80]
  (T.east) -- (yt.west);

\draw[-{Triangle[length=1.8mm,width=1.5mm]}, thick, draw=black!80]
  (S.east) -- (ys.west);

\end{tikzpicture}
    
    \caption{Overview of the \method framework. 
    Each context $\mathbf{c}_i$ is mapped into an adapter by the generator $G$, while the query $\mathbf{q}$ is processed by the student $S$. The teacher $T$ processes the concatenated input $[\mathbf{c};\mathbf{q}]$, and $G$ is trained so that composed adapters in $S$ align $\mathbf{y}_s$ with $\mathbf{y}_t$, effectively transferring contextual information from input space to parameter space.}
    \label{fig:compas_overview}
\end{figure*}
These works largely focused on producing adapters from a single \textit{context}.
In practice, however, contexts rarely occur in isolation. Tasks are often guided by chains of instructions, multiple demonstrations, or sets of retrieved passages that must be integrated.  
Without a principled notion of composition, such adapters cannot faithfully capture how LLMs combine multiple sources of context.  
Composability also addresses the limitations of prompt-based adaptation. Instead of concatenating long contexts, adapters can be merged algebraically in parameter space. The key challenge is therefore not parameter generation alone, but algebraically consistent composition that preserves semantics of concatenated texts while retaining efficient inference.  

To bridge this critical gap, we introduce \method (\textbf{Comp}ositional \textbf{A}dapter\textbf{s}), a framework that maps context into adapter parameters with compositional structure at its core. See \Cref{fig:compas_overview} for a high-level overview.
In \method, a teacher LM process concatenated contexts, while the student LM learns to approximate this behavior using only query tokens, complemented with the sum of adapters generated for individual contexts.
This is facilitated by a generator network, which maps individual contexts into compositional adapters. 
\method is driven by auxiliary compositionality and reconstruction losses, which provide diagnostics for whether parameter-space operations preserve semantics and faithfulness \citep{jacovi2020towards, turpin2023language, lampinen2022can}. 
We evaluate \method 
on multiple-choice and extractive QA tasks, where either demonstrations (MMLU, ARC-Challenge) or retrieved passages (HotpotQA) serve as context.
We observe consistent improvements over ICL, fine-tuning and existing context-to-parameter methods. These gains are most pronounced when scaling to larger numbers of demonstrations, highlighting the capacity of \method for efficient integration of external evidence by composition in parameter-space.

Our contributions are threefold: (1) We introduce \method, a teacher-student framework for encoding context into adapters which facilitates composition in parameter space. (2) We establish theoretical conditions under which parameter-space addition provably approximates the behavior of the teacher model. (3) We empirically show that \method outperforms alternatives, especially when scaling to large context sizes, demonstrating the benefits of compositional integration of information.

\section{Adapter Compositionality}
\label{sec:theory}

Our goal is to ensure that adapters generated from individual contexts can be \emph{composed} in parameter space to replicate the effect of context concatenation in input space. More concretely, in a teacher--student setup, the outputs of the teacher LM produced when given multiple contexts jointly should be reproduced by the student LM using only the query tokens along with the sum of the corresponding adapter parameters.
We now formalize this requirement, introducing conditions under which adapter addition in the parameter space corresponds to context concatenation in the input space.

\subsection{Setup}

Let $V$ be the vocabulary and $|V|$ its size. A language model with frozen parameters 
$\boldsymbol{\theta} \in \mathbb{R}^n$ is a function
$
\mathbf{f}_{\boldsymbol{\theta}} : \mathcal{X} \to \mathbb{R}^{|V|},
$
where $\mathcal{X} = \Sigma^*$ is the set of all token sequences over the alphabet $\Sigma$. 
For $\mathbf{x} \in \mathcal{X}$, the output $\mathbf{f}_{\boldsymbol{\theta}}(x) \in \mathbb{R}^{|V|}$ 
denotes the logit vector over the vocabulary.
A \emph{support context} $\mathbf{c} \in \mathcal{C}$ is any textual information 
(e.g., a demonstration, instruction, or retrieved passage) provided alongside a query 
$\mathbf{q} \in \mathcal{X}$, where $\mathcal{C} \subseteq \mathcal{X}$ denotes the 
set of admissible contexts. 
We use $[\mathbf{c};\mathbf{q}] \in \mathcal{X}$ to denote the concatenation of a 
context $\mathbf{c}$ with the query $\mathbf{q}$.

Let $\Phi \subseteq \mathbb{R}^m$ denote the set of adapter parameters, and let $\boldsymbol{\phi} \in \Phi$ denote parameters for a single adapter, i.e., a structured modification of $\boldsymbol{\theta}$.  
We write $\boldsymbol{\theta} \oplus \boldsymbol{\phi}$ for the parameters obtained by composing $\boldsymbol{\theta}$ with $\boldsymbol{\phi}$.

We now formalize the teacher--student setup that underlies our framework. The \textbf{teacher} model corresponds to the base language model applied to the concatenated input:
\[
\mathbf{f}_T(\mathbf{c},\mathbf{q})
\coloneqq
\mathbf{f}_{\boldsymbol{\theta}}\!\left([\mathbf{c};\mathbf{q}]\right).
\]

The \textbf{student} model processes only the query, using parameters modified by an adapter:
\[
\mathbf{f}_S^{\boldsymbol{\phi}}(\mathbf{q})
\coloneqq
\mathbf{f}_{\boldsymbol{\theta}\oplus\boldsymbol{\phi}}(\mathbf{q}).
\]
Throughout this work, both teacher and student share the same frozen base 
parameters $\boldsymbol{\theta}$, ensuring that any behavioral differences arise 
solely from the presence or absence of adapter-based context encoding. While the 
formulation could accommodate distinct base models in principle, we deliberately keep the underlying model fixed to isolate the effect of adapters.

For a sequence of contexts $\mathbf{c}_1,\ldots,\mathbf{c}_k$ (denoted compactly by $\mathbf{c}_{1:k}$), the teacher receives the concatenated input $[\mathbf{c}_{1:k};\mathbf{q}]$, while the student processes only $\mathbf{q}$ together with the summed adapter parameters $\sum_{i=1}^k \boldsymbol{\phi}_i$. For example, with two contexts $\mathbf{c}_1$ and $\mathbf{c}_2$, the student model becomes
\[
\mathbf{f}_S^{\boldsymbol{\phi}_1+\boldsymbol{\phi}_2}(\mathbf{q})
=
\mathbf{f}_{\boldsymbol{\theta} \oplus (\boldsymbol{\phi}_1+\boldsymbol{\phi}_2)}(\mathbf{q}).
\]

Finally, a \textbf{generator} $G:\mathcal{C}\to\Phi$ maps each context $\mathbf{c}\in\mathcal{C}$ to an adapter $\boldsymbol{\phi}=G(\mathbf{c})$.  
In particular, for two contexts $\mathbf{c}_1$ and $\mathbf{c}_2$, we define
\[
\boldsymbol{\phi}_1 = G(\mathbf{c}_1),
\quad
\boldsymbol{\phi}_2 = G(\mathbf{c}_2),
\quad
\boldsymbol{\phi}_{12} = G([\mathbf{c}_1;\mathbf{c}_2]).
\]


\subsection{Compositionality Bound}

The set of finite contexts $\mathcal{C}$, equipped with concatenation and the empty string identity ($\lambda$), forms a free monoid $(\mathcal{C},[\cdot;\cdot],\lambda)$. On the other hand, the adapter space $(\Phi,+,\mathbf{0})$ is a commutative additive monoid.  
A fundamental requirement for adapter compositionality is that the generator $G:\mathcal{C}\to\Phi$ is a monoid homomorphism,
$
G([\mathbf{c}_1;\mathbf{c}_2]) = G(\mathbf{c}_1)+G(\mathbf{c}_2)$
and
$
G(\lambda)=\mathbf{0}.
$
This ensures that addition in the adapter space exactly mirrors concatenation in the input space.
Demanding $G$ to be a homomorphism aligns these two structures.
Concatenation on $\mathcal{C}$ is associative but not commutative, whereas addition on $\Phi$ is both.
Thus, requiring $G$ to be a homomorphism implicitly collapses the order of contexts.
This means that $G$ is a non-injective homomorphism: permuted concatenations map to the same adapter sum. We regard this loss of order as desirable in many cases, as the contribution of a demonstration or retrieved passage should often be independent of its relative position. By treating contexts as a set rather than a sequence, we aim for adapters that focus on semantics, mitigating the order sensitivity and instability observed in long-context transformers.%
\footnote{Certain tasks may still require positional information (e.g., reasoning with ordered 
evidence). We show in Appendix~\ref{sec:position} that our framework can incorporate 
explicit positional encodings when needed.}

Enforcing a homomorphism between the free monoid of contexts and the additive monoid of adapter parameters via a teacher--student setup cannot be achieved exactly in practice, given the models' finite representational capacity, the finiteness of the training sample, and the approximate nature of optimization. 
We therefore approximate it, which introduces 
discrepancies. To capture these discrepancies and regularity conditions, we introduce three quantities.

\paragraph{Student--teacher error.}  
For a context $\mathbf{c}$ with the corresponding adapter $\boldsymbol{\phi}=G(\mathbf{c})$, the student’s discrepancy from the teacher is  
\begin{equation}
\epsilon(\mathbf{c}) := 
\|\mathbf{f}_S^{\boldsymbol{\phi}}(\mathbf{q}) 
   - \mathbf{f}_T([\mathbf{c};\mathbf{q}])\|_2,
\label{eq:student_teacher_error}
\end{equation}

\paragraph{Generator additivity error.}  
The generator’s deviation from additivity is  
\begin{equation}
\eta := \|G([\mathbf{c}_1;\mathbf{c}_2]) - (G(\mathbf{c}_1)+G(\mathbf{c}_2))\|_2.
\label{eq:generator_additivity_error}
\end{equation}

\paragraph{Parameter sensitivity.}  
The student’s logits vary smoothly with respect to adapter parameters, with a Lipschitz constant  
\begin{equation}
L := \sup_{\boldsymbol{\phi}_1,\boldsymbol{\phi}_2,\mathbf{q}}
\frac{\|\mathbf{f}_S^{\boldsymbol{\phi}_1}(\mathbf{q}) - \mathbf{f}_S^{\boldsymbol{\phi}_2}(\mathbf{q})\|_2}
     {\|\boldsymbol{\phi}_1-\boldsymbol{\phi}_2\|_2}.
\label{eq:parameter_sensitivity}
\end{equation}




The first two quantities capture the core requirements of compositionality, while the third 
provides a regularity condition needed for the analysis.  Assuming $L < \infty$, i.e., Lipschitz continuity of the student with respect to adapter parameters, we obtain the following result.
\begin{theorem}[Compositionality Bound]
\label{thm:compositionality}
Let $\boldsymbol{\phi} = G(\mathbf{c})$ denote the adapter parameters generated from a context $\mathbf{c}$. 
Let $\epsilon$ denote the teacher--student compositional error, $\eta$ the generator additivity error, and $L$ the sensitivity of the student with respect to adapter parameters. 
Then, for any contexts $\mathbf{c}_1,\mathbf{c}_2$ and query $\mathbf{q}$,
\[
\|\mathbf{f}_S^{\boldsymbol{\phi}_1+\boldsymbol{\phi}_2}(\mathbf{q})
 - \mathbf{f}_T([\mathbf{c}_1;\mathbf{c}_2;\mathbf{q}])\|_2
\;\le\; L \eta \;+\; \epsilon([\mathbf{c}_1;\mathbf{c}_2]).
\]
\end{theorem}
See Appendix~\ref{app:proof} for the proof. 
The bound decomposes the student--teacher error into two interpretable sources: generator additivity error $\eta$ and misfit on the concatenated context $\epsilon([\mathbf{c}_1;\mathbf{c}_2])$.  
The Lipschitz constant $L$ propagates generator errors into the student. Thus, perfect compositionality is unattainable in practice, but approximate compositionality is achievable whenever these quantities remain small. 

\section{Method}
\label{sec:method}

Previously, we decomposed the discrepancy between \emph{context concatenation} and \emph{adapter addition} into two main error sources: generator additivity error ($\eta$) and student--teacher error on concatenated contexts ($\epsilon$), with parameter sensitivity ($L$) controlling how generator errors propagate to outputs.  
While $L$ is a structural property of the student model, we can design the generator and its training objectives to directly minimize $\eta$ and $\epsilon$. To this end, we introduce \method, a teacher--student meta-learning framework that maps each support context into a LoRA adapter \citep{hu2022lora}, which is composed additively with the base parameters. The adapter generator ends with a linear bottleneck, ensuring that its outputs in parameter space add directly.
This algebraic addition is trained to approximate the \emph{outputs} of the model under context concatenation in input space. To achieve this, we design loss functions that incentivize the generator to approximate a homomorphism from context concatenation to parameter-space addition.

\subsection{Context-to-Adapter Generator}
\label{sec:generator}
 
As introduced in \Cref{sec:theory}, the generator $G$ maps a context $\mathbf{c}$ to adapter parameters $\boldsymbol{\phi}=G(\mathbf{c})$. 
It is implemented by augmenting the base LM with additional trainable components:
(i) its own LoRA adapter inserted into the LM, and (ii) a linear bottleneck followed 
by two projection trunks. The same frozen base LM is shared across the teacher, the 
student, and the generator, where only the generator’s adapter and projection components 
are updated during training.

Given a support context $\mathbf{c}$, the base LM (with the generator’s adapter) 
processes the tokens, and we obtain
$
\mathbf{h}(\mathbf{c}) \in \mathbb{R}^d
$
as the mean of the final-layer hidden states across all tokens.  
This pooled representation is projected into a compact latent space,
\[
\mathbf{z}(\mathbf{c}) = P\,\mathbf{h}(\mathbf{c}), 
\qquad P \in \mathbb{R}^{r\times d}, \;\; r \ll d,
\]
reducing dimensionality and ensuring the subsequent mapping into LoRA parameters remains tractable. Without this bottleneck, a direct mapping would require prohibitively many parameters.

A LoRA module applied to a target linear map with input dimension $d^{\mathrm{in}}$ 
and output dimension $d^{\mathrm{out}}$ consists of two low-rank matrices
$
\mathbf{A} \in \mathbb{R}^{r \times d^{\mathrm{in}}} 
$
and
$
\mathbf{B} \in \mathbb{R}^{d^{\mathrm{out}} \times r},
$
with rank $r \ll d^{\mathrm{in}},d^{\mathrm{out}}$.  
Each transformer layer may contain multiple such modules (e.g., query, key, value, 
or output projections). Let $\mathcal{M}$ denote the set of all modules instrumented 
with LoRA, indexed across layers and projection types.
From the latent representation $\mathbf{z}(\mathbf{c})$, the generator produces parameters for all modules via two bias-free linear projections:
$
U_A \mathbf{z}(\mathbf{c}) \in \mathbb{R}^{m_A}
$
and
$
U_B \mathbf{z}(\mathbf{c}) \in \mathbb{R}^{m_B},
$
with
\[
m_A = \sum_{m \in \mathcal{M}} r_m d_m^{\mathrm{in}}, 
\qquad 
m_B = \sum_{m \in \mathcal{M}} d_m^{\mathrm{out}} r_m.
\]
The outputs of $U_A$ and $U_B$ are partitioned and reshaped into the individual 
LoRA matrices $\mathbf{A}_m(\mathbf{c})$ and $\mathbf{B}_m(\mathbf{c})$ for each 
target module $m \in \mathcal{M}$. See \Cref{fig:compas_overview} for a high-level overview of \method.

\subsection{Loss Function Components}
\label{sec:losses}

Our loss function design follows Theorem~\ref{thm:compositionality}: we reduce the student--teacher error in both single and concatenated contexts, and penalize generator non-additivity. We also include a reconstruction term that encourages adapters to faithfully encode their contexts (up to permutation), preventing collapse into trivial solutions. This auxiliary objective also regularizes training and provides a stronger learning signal, aiding optimization.

\paragraph{Student--teacher alignment.}
We draw unlabeled queries $\mathbf{q}$ together with $k$ contexts, and obtain 
soft pseudo-labels from the teacher.
For brevity, we introduce the shorthands
$
\mathbf{p}_T(\mathbf{c}_{1:k},\mathbf{q})
$
and
$
\mathbf{p}_S^{\boldsymbol{\phi}}(\mathbf{q})
$,
denoting the teacher and student predictive distributions, respectively.
Concretely, we define
\begin{align*}
\mathbf{p}_T(\mathbf{c}_{1:k},\mathbf{q})
& = \mathrm{softmax}\!\big(
\mathbf{f}_T([\mathbf{c}_{1:k};\mathbf{q}])
\big), \\
\mathbf{p}_S^{\boldsymbol{\phi}}(\mathbf{q})
& = \mathrm{softmax}\!\big(
\mathbf{f}_S^{\sum_{i=1}^k \boldsymbol{\phi}_i}(\mathbf{q})
\big).
\end{align*}

The student is trained to match the teacher’s predictive distribution via \textit{weak supervision} by minimizing the Kullback--Leibler divergence, denoted by $D_{\mathrm{KL}}(\cdot \,\|\, \cdot)$, between their output distributions:
\begin{align*}
\mathcal{L}_{\mathrm{ST}}
= \mathbb{E}_{(\mathbf{c}_1,\ldots,\mathbf{c}_k),\,\mathbf{q}}
\bigg[
D_{\mathrm{KL}}
\!\Big(
\mathbf{p}_T(\mathbf{c}_{1:k},\mathbf{q})
\;\big\|\;
\mathbf{p}_S^{\boldsymbol{\phi}}(\mathbf{q})
\Big)
\bigg].
\end{align*}
Our overall objective considers $k \in \{1,2\}$, where $k=1$ enforces fidelity to a single context, while $k=2$ encourages compositional consistency across two contexts.

\paragraph{Additivity regularization.}
We reduce the generator additivity error $\eta$ by penalizing discrepancies between parameters generated for concatenated contexts and the sum of parameters from 
the individual contexts. Concretely, we draw pairs of contexts $(\mathbf{c}_1,\mathbf{c}_2)$ and compare generated parameters. For each module $m \in \mathcal{M}$, let 
$\mathbf{W}_m(\mathbf{c}) \in \{\mathbf{A}_m(\mathbf{c}), \mathbf{B}_m(\mathbf{c})\}$
denote one of the two LoRA matrices generated from context $\mathbf{c}$. 
We write $\|\cdot\|_F$ for the Frobenius norm and let $\delta>0$ be a small stability constant. 
We define the normalized additivity discrepancy
\[
\Delta_m(\mathbf{c}_1,\mathbf{c}_2)
=
\frac{
\left\|
\mathbf{W}_m([\mathbf{c}_{1:2}])
-
\mathbf{W}_m(\mathbf{c}_1)
-
\mathbf{W}_m(\mathbf{c}_2)
\right\|_F^2
}{
\left\|
\mathbf{W}_m([\mathbf{c}_{1:2}])
\right\|_F^2
+ \delta
},
\]
and minimize the additivity objective
\begin{equation*}
\mathcal{L}_{\mathrm{ADD}}
=
\mathbb{E}_{(\mathbf{c}_1,\mathbf{c}_2)}
\left[
\sum_{m \in \mathcal{M}}
\Delta_m(\mathbf{c}_1,\mathbf{c}_2)
\right].
\end{equation*}

\paragraph{Reconstruction.}
To encourage faithfulness and improve training stability, we require adapters to recoverably encode the information contained in their contexts. 
To this end, we introduce a special query token \texttt{[RECON]}, which prompts the student to reconstruct the support context $\mathbf{c} = (c_1,\ldots,c_T)$ autoregressively from its adapter state, where $c_t$ denotes the token at position $t$ and $\mathbf{c}_{<t} = (c_1,\ldots,c_{t-1})$ the prefix preceding position $t$. 
Let $p_S^{\boldsymbol{\phi}}(v \mid \mathbf{s})$ denote the student’s conditional predictive distribution over the next token $v$ given an arbitrary conditioning sequence $\mathbf{s}$. Under the reconstruction prompt, the conditioning sequence is $\mathbf{s} = \left[\texttt{[RECON]}; \mathbf{c}_{<t}\right]$. This discourages collapse to trivial solutions and provides an auxiliary learning signal through the cross-entropy loss:
\begin{equation*}
\mathcal{L}_{\mathrm{RECON}}
=
- \mathbb{E}_{\mathbf{c}}
\left[
\sum_{t=1}^{T}
\log p_S^{\boldsymbol{\phi}}(c_t \mid \texttt{[RECON]}; \mathbf{c}_{<t})
\right].
\end{equation*}

\paragraph{Overall objective.}
The final loss function is a weighted sum:
\begin{equation}
\mathcal{L}_{\method}
=\; \lambda_{\text{ST}}\mathcal{L}_{\text{ST}}
+\lambda_{\text{ADD}}\mathcal{L}_{\text{ADD}}
+\lambda_{\text{RECON}}\mathcal{L}_{\text{RECON}}.
\label{eq:objective}
\end{equation}
The coefficients $\lambda_{\text{ST}}, \lambda_{\text{ADD}},$ and $\lambda_{\text{RECON}}$ balance alignment, additivity regularization, and reconstruction fidelity, respectively. 
Together, these terms support compositional generalization while maintaining faithful encoding of contextual information.
Hyperparameter choices and training details are provided in Appendix~\ref{app:hyper}.

\section{Experiments}
\label{sec:experiments}

We first outline the experimental design \Cref{sec:setup}, with full setup and hyperparameter details in Appendix~\ref{app:hyper}. We then evaluate \method in two regimes: (i) replacing in-context demonstrations with adapter parameters (\Cref{sec:icl}), and (ii) encoding retrieved passages as parametric memory (\Cref{sec:qa}). Finally, we assess reconstruction, measuring faithfulness of information encoded in adapter parameters (\Cref{sec:recon}). Additional experiments on context order sensitivity and efficiency are in Appendix~\ref{app:experiments}.

\subsection{Experimental Setup}
\label{sec:setup}

\paragraph{Models.}
We experiment with decoder-only LMs: \textbf{LLaMA-3.1~8B} (LLaMA 8B) and \textbf{Llama-3.2~3B} \citep{llama3}, and \textbf{Qwen-2.5~7B}  \citep{qwen2025}. For brevity, we refer to these models as LLaMA 8B, LLaMA 3B, and Qwen 7B.

\paragraph{Tasks.}
We consider two representative settings:  
(i) ICL on \textbf{MMLU} \citep{hendrycks2021mmlu} and \textbf{ARC-Challenge} \citep{clark2018arc}, where few-shot exemplars are provided as demonstrations;  
(ii) Extractive question answering (QA) on \textbf{HotpotQA} \citep{yang2018hotpotqa}, where gold passages supply contextual evidence.  
Prompts and preprocessing are standardized across all methods (\Cref{app:prompts}).  

\paragraph{Methods.} 
We compare against three representative approaches. 
\textbf{ICL} is standard $k$-shot prompting with demonstrations concatenated in prompt text. 
\textbf{Generative Adapter (GenAda)} \citep{chen2025generativeadapter} employs a nonlinear hypernetwork to predict adapter weights from context, trained with reconstruction and continuation losses. In contrast, our method enforces a linear bottleneck, generating adapters via a lightweight adapter module followed by a linear transformation. 
\textbf{\wilda} \citep{jukic-2025-wilda} fine-tunes a separate adapter for each context, achieving strong accuracy but incurring heavy computational overhead, since a new adapter must be trained for every context.


\paragraph{Evaluation.}
We assess models on end-task performance, measuring accuracy on MMLU and ARC and exact match (EM) and token-level F1 on HotpotQA. Faithfulness is evaluated through the KL divergence between the teacher concatenation and the student sum, as well as token-level F1 (\textit{bag-of-tokens}). Stability is measured as the standard deviation across ten runs with different sampled contexts. Training follows the weakly supervised protocol from \Cref{sec:losses}. Full implementation details, including optimization settings and hyperparameters, are deferred to \Cref{app:hyper}.

\subsection{Encoding Demonstrations as Parameters}
\label{sec:icl}
We first evaluate effectiveness in replacing in-context demonstrations with adapter parameters. In the $k$-shot setting ($k \in \{4,8,12,16\}$), demonstrations are partitioned into fixed-size blocks, each block is encoded as an adapter, and the adapters are composed by summing their parameters. \Cref{tab:icl} shows the results on MMLU and ARC-Challenge across three base models.

\begin{table*}[t]
\small
\centering
\caption{Demonstration parameterization results. Rows with a single number indicate concatenation of all demonstrations without composition, while rows of the form $a{\times}b$ denote composition of $a$ adapters, each encoding $b$ demonstrations. Results are reported as the mean over $10$ runs with the standard deviation as a subscript. The best score within each block is highlighted in \textbf{bold}. \method results are statistically compared to the corresponding ICL setting; scores marked with $\dagger$ indicate significance under a Wilcoxon signed-rank test ($p<0.05$) with Holm–Bonferroni correction.}
\begin{tabular}{l cc  cc  cc}
\toprule
& \multicolumn{2}{c}{\textbf{Llama-3.1 8B}} & \multicolumn{2}{c}{\textbf{Llama-3.2 3B}} & \multicolumn{2}{c}{\textbf{Qwen-2.5 7B}} \\
\cmidrule(lr){2-3}\cmidrule(lr){4-5}\cmidrule(lr){6-7}
\textbf{Setup / Method} & \textbf{MMLU} & \textbf{ARC} & \textbf{MMLU} & \textbf{ARC} & \textbf{MMLU} & \textbf{ARC} \\
\midrule
4 \quad ICL                     & $64.2_{1.8}$ & $74.2_{1.6}$ & $57.2_{1.8}$ & $62.9_{1.6}$ & $70.2_{1.8}$ & $76.5_{1.6}$ \\
4 \quad GenAda      & $57.6_{1.8}$ & $69.3_{1.7}$ & $50.4_{1.4}$ & $59.4_{1.7}$ & $59.8_{2.1}$ & $70.5_{1.7}$ \\
4 \quad \wilda                   & $\mathbf{68.8}_{0.9}$ & $\mathbf{78.2}_{0.7}$ & $\mathbf{60.8}_{1.1}$ & $\mathbf{66.4}_{1.2}$ & $\mathbf{74.0}_{0.4}$ & $\mathbf{79.6}_{0.9}$ \\
4 \quad \method        & $66.7_{0.7}$\nospacetext{$^\dagger$} & $76.5_{0.8}$\nospacetext{$^\dagger$} & $59.0_{0.9}$ & $65.1_{0.8}$\nospacetext{$^\dagger$} & $72.4_{0.5}$\nospacetext{$^\dagger$} & $77.2_{0.6}$ \\
\midrule
8 \quad ICL                     & $65.5_{1.7}$ & $75.1_{1.5}$ & $58.0_{1.7}$ & $64.1_{1.5}$ & $71.5_{1.7}$ & $78.3_{1.5}$ \\
8 \quad GenAda     & $59.1_{1.9}$ & $69.9_{1.6}$ & $51.3_{1.9}$ & $61.2_{1.6}$ & $60.7_{1.9}$ & $71.4_{1.6}$ \\ 
8 \quad \wilda                   & $\mathbf{69.8}_{0.7}$ & $78.8_{0.8}$ & $61.2_{1.1}$ & $\mathbf{67.4}_{0.5}$ & $\mathbf{74.8}_{0.7}$ & $80.0_{0.8}$ \\
8 \quad \method         & $67.5_{0.5}$ & $77.2_{0.9}$\nospacetext{$^\dagger$} & $60.2_{1.3}$\nospacetext{$^\dagger$} & $66.2_{0.5}$\nospacetext{$^\dagger$} & $73.2_{1.2}$ & $78.6_{0.7}$ \\
\cmidrule(lr){1-1}
$2{\times}4$ \, GenAda  & $57.8_{1.9}$ & $69.5_{1.6}$ & $50.9_{0.7}$ & $59.4_{1.2}$ & $60.5_{1.8}$ & $69.2_{1.6}$ \\   
$2{\times}4$ \, \wilda           & $69.5_{0.3}$  & $78.6_{0.9}$ & $60.8_{1.1}$ & $67.1_{1.0}$ & $74.1_{0.4}$ & $79.8_{0.8}$ \\
$2{\times}4$ \, \textsc{CompAS} & $69.1_{0.7}$\nospacetext{$^\dagger$} & $\mathbf{79.3}_{0.9}$\nospacetext{$^\dagger$} & $\mathbf{61.3}_{1.2}$\nospacetext{$^\dagger$} & $66.8_{0.6}$\nospacetext{$^\dagger$} & $74.6_{0.6}$\nospacetext{$^\dagger$} & $\mathbf{80.2}_{0.9}$ \\
\midrule
12 \quad ICL                    & $66.0_{1.7}$ & $75.5_{1.5}$ & $58.8_{1.7}$ & $64.6_{1.5}$ & $71.8_{1.7}$ & $77.2_{1.5}$ \\
12 \quad GenAda     & $59.9_{1.9}$ & $70.2_{1.6}$ & $51.6_{1.9}$ & $62.5_{1.6}$ & $63.1_{1.9}$ & $71.6_{1.2}$ \\
12 \quad \wilda                  & $70.5_{0.8}$ & $79.4_{0.6}$ & $61.9_{1.1}$ & $\mathbf{68.1}_{1.0}$ & $75.2_{0.7}$ & $79.8_{0.4}$ \\
12 \quad \method        & $67.2_{1.2}$ & $78.9_{0.9}$\nospacetext{$^\dagger$} & $59.5_{1.2}$ & $67.2_{0.7}$\nospacetext{$^\dagger$} & $72.9_{1.0}$ & $79.3_{1.4}$ \\
\cmidrule(lr){1-1}
$3{\times}4$ \, GenAda      & $57.3_{1.9}$ & $70.3_{1.6}$ & $49.8_{1.9}$ & $59.6_{1.6}$ & $62.0_{1.9}$ & $69.7_{1.6}$ \\
$3{\times}4$ \, \wilda           & $70.7_{1.0}$ & $79.5_{0.9}$ & $62.0_{1.1}$ & $67.5_{1.0}$ & $75.1_{0.7}$ & $79.9_{0.4}$ \\
$3{\times}4$ \, \method & $\mathbf{71.2}_{1.1}$\nospacetext{$^\dagger$} & $\mathbf{80.1}_{0.5}$\nospacetext{$^\dagger$} & $\mathbf{62.4}_{0.8}$\nospacetext{$^\dagger$} & $67.4_{0.9}$ & $\mathbf{75.7}_{0.3}$\nospacetext{$^\dagger$} & $\mathbf{80.3}_{0.8}$\nospacetext{$^\dagger$} \\
\midrule
16 \quad ICL                    & $66.5_{1.7}$ & $76.0_{1.5}$ & $59.2_{1.7}$ & $65.2_{1.5}$ & $72.3_{1.7}$ & $77.6_{1.5}$ \\
16 \quad GenAda     & $60.5_{1.9}$ & $72.5_{1.6}$ & $52.7_{1.2}$ & $60.8_{1.3}$ & $64.2_{1.4}$ & $71.9_{0.9}$ \\
16 \quad \wilda                  & $71.5_{0.7}$ & $80.4_{0.3}$ & $62.6_{0.5}$ & $68.2_{0.3}$ & $76.1_{0.4}$ & $80.7_{0.8}$ \\
16 \quad \method        & $68.2_{0.6}$ & $79.1_{0.7}$\nospacetext{$^\dagger$} & $62.3_{1.2}$ & $68.3_{0.4}$\nospacetext{$^\dagger$} & $73.7_{0.6}$ & $79.4_{0.5}$ \\
\cmidrule(lr){1-1}
$4{\times}4$ \,  GenAda      & $57.6_{1.9}$ & $70.6_{1.6}$ & $50.3_{1.9}$ & $59.9_{1.6}$ & $61.3_{1.9}$ & $70.4_{1.6}$ \\
$4{\times}4$ \, \wilda           & $71.6_{0.6}$ & $80.5_{0.9}$ & $62.7_{1.1}$ & $68.3_{1.1}$ & $75.2_{0.5}$ & $80.8_{0.8}$ \\
$4{\times}4$ \, \method & $\mathbf{72.2}_{0.3}$\nospacetext{$^\dagger$} & $\mathbf{81.3}_{0.3}$\nospacetext{$^\dagger$} & $\mathbf{63.4}_{0.7}$\nospacetext{$^\dagger$} & $\mathbf{69.3}_{0.5}$\nospacetext{$^\dagger$} & $\mathbf{77.1}_{0.7}$\nospacetext{$^\dagger$} & $\mathbf{81.5}_{0.4}$\nospacetext{$^\dagger$} \\
\bottomrule
\end{tabular}
\label{tab:icl}
\end{table*}

\textit{\method consistently outperforms standard ICL}, which in our setup corresponds to using the teacher model directly, across all models and shot counts, while also exhibiting lower variance. We attribute this robustness to weak-to-strong (W2S) generalization \citep{dherin-etal-2022-neural,lang-etal-2024-theoretical}: the student begins with weak pseudo-labels from the teacher but progressively corrects them during training, extending reliability from easy, locally consistent regions to harder examples.

Gains are most pronounced at higher shot counts ($k \in {12,16}$) under $3{\times}4$ and $4{\times}4$ composition settings, where \method consistently outperforms \wilda (11 out of 12 cases), despite the latter’s use of context-specific adapter fine-tuning. We hypothesize that this advantage arises from an implicit chunking--composition mechanism, where the model partitions demonstrations into manageable subsets. Each adapter encodes a partial context view, and their addition reconstructs the effect of all demonstrations. As a result, \method scales gracefully with the number of demonstrations: it leverages composition to maintain strong performance in long contexts, ultimately matching or exceeding specialized adapter fine-tuning while retaining efficiency and composability. A more detailed analysis of how block granularity affects compositional stability and scaling behavior is provided in Appendix~\ref{sec:comp_scaling}.

\subsection{Encoding Context as Parametric Memory}
\label{sec:qa}

We now evaluate \method on extractive QA by treating gold evidence passages as \emph{parametric memories}.  
On HotpotQA, each query $\mathbf{q}$ is paired with one or more gold passages $\mathbf{c}$.
When multiple passages are available (pairs or triplets), the task becomes more challenging: the model must leverage a larger combined context and answer more queries. In these settings, we either generate a single adapter from their concatenation or compose the adapters generated from each passage individually. We evaluate per query, including all gold passages in either form. Results are shown in \Cref{tab:extractive}.

\method consistently outperforms the teacher across all settings.  
Absolute performance is lower on pairs and triplets -- since the model must retain context for multiple queries while being evaluated on each one separately -- but the relative gains over the teacher are larger in these harder settings. This indicates that \method is particularly effective at composing and retaining information across multiple contexts.

\begin{table*}[t]
\small
\centering
\caption{Extractive QA results on single, pair, and triplet gold contexts, averaged over $10$ runs and reported as EM/F1. \emph{Teacher (concat)} evaluates the base LLM directly on concatenated gold contexts (e.g., $[\mathbf{c}_1;\mathbf{c}_2;\mathbf{c}_3]$). 
\method~(\emph{concat}) evaluates the student using an adapter generated from the concatenated context, 
$G([\mathbf{c}_1;\mathbf{c}_2;\mathbf{c}_3])$, 
while \method~(\emph{composed}) uses the sum of adapters from individual contexts, $\sum_i G(\mathbf{c}_i)$.}
\label{tab:extractive}
\begin{tabular}{llccc}
\toprule
\textbf{Setup} & \textbf{Method} & \textbf{LLaMA-3.1 8B} & \textbf{LLaMA-3.2 3B} & \textbf{Qwen-2.5 7B} \\
\midrule
\multirow{2}{*}{\textbf{Single ($\mathbf{c}$)}} 
& Teacher  & $70.9 / 82.0$ & $65.9 / 78.1$ & $69.2 / 80.7$ \\  
& \textsc{CompAS} & $72.2 / 82.9$ & $67.0 / 78.6$ & $69.4 / 82.2$ \\
\midrule
\multirow{3}{*}{\textbf{Pair ($[\mathbf{c}_1;\mathbf{c}_2]$)}} 
& Teacher (concat)              & $69.0 / 80.8$ & $64.1 / 76.8$ & $67.5 / 79.8$ \\ 
& \textsc{CompAS} (concat) & $69.4 / 81.1$ & $64.5 / 77.0$ & $67.9 / 79.5$ \\
& \textsc{CompAS} (composed) & $71.3 / 82.3$ & $66.2 / 78.4$ & $69.6 / 80.9$ \\
\midrule
\multirow{3}{*}{\textbf{Triplet ($[\mathbf{c}_1;\mathbf{c}_2;\mathbf{c}_3]$)}} 
& Teacher (concat) & $66.3 / 78.5$ & $61.2 / 74.1$ & $64.0 / 76.0$ \\             
& \textsc{CompAS} (concat) & $67.5 / 79.4$ & $62.0 / 75.3$ & $65.1 / 77.0$ \\
& \textsc{CompAS} (composed)  & $71.1 / 82.0$ & $64.8 / 76.9$ & $68.1 / 80.4$ \\
\bottomrule
\end{tabular}
\end{table*}

\subsection{Context Reconstruction}
\label{sec:recon}

Finally, we test whether composed adapters preserve the informational content of their supports by prompting the model to reconstruct missing contexts.  
Using a special \texttt{[RECON]} token, we decode from adapters and compute token-level F1 on MMLU, ARC, and HotpotQA.  
\Cref{tab:compas_reconstruction} shows that \method maintains consistently high reconstruction fidelity, with only small drops as the context expands to pair and triplet units. In \Cref{sec:loss-analysis}, we further show that ablating the reconstruction objective leads to a consistent drop in downstream performance, highlighting its importance beyond faithfulness alone.

\begin{table*}[]
\small
\centering
\caption{\method context reconstruction as token-level F1. Results are averages over $10$ runs.
MMLU and ARC \emph{units} are blocks of $4$ demonstrations; a HotpotQA \emph{unit} is a supporting paragraph. }
\begin{tabular}{l ccc ccc ccc}
\toprule
& \multicolumn{3}{c}{\textbf{Llama-3.1 8B}} & \multicolumn{3}{c}{\textbf{Llama-3.2 3B}} & \multicolumn{3}{c}{\textbf{Qwen-2.5 7B}} \\
\cmidrule(lr){2-4}\cmidrule(lr){5-7}\cmidrule(lr){8-10}
\textbf{Setup} & \textbf{MMLU} & \textbf{ARC} & \textbf{Hotpot} & \textbf{MMLU} & \textbf{ARC} & \textbf{Hotpot} & \textbf{MMLU} & \textbf{ARC} & \textbf{Hotpot} \\
\midrule
Single (1 unit)   & $91.2$ & $89.5$ & $87.8$ & $89.0$ & $87.2$ & $85.4$ & $90.1$ & $88.1$ & $86.2$ \\
Pair (2 units)    & $90.3$ & $88.9$ & $86.0$ & $88.1$ & $86.3$ & $83.5$ & $89.4$ & $87.5$ & $84.3$ \\
Triplet (3 units) & $88.4$ & $86.9$ & $83.7$ & $86.5$ & $84.8$ & $81.9$ & $87.6$ & $85.9$ & $82.5$ \\
\bottomrule
\end{tabular}
\label{tab:compas_reconstruction}
\end{table*}

Beyond accuracy, compositionality, and reconstruction fidelity, we also examine the computational footprint of \method. A detailed efficiency analysis is provided in Appendix~\ref{sec:efficiency}, where we compare inference FLOPs, peak memory, and cumulative training cost across methods. The results show that \method decouples inference cost from context length, yielding increasing speedups at higher shot counts while maintaining a moderate amortized training cost.

\section{Analysis}
\label{sec:analysis}

Building on the effectiveness of \method, we analyze the role of generator capacity (\Cref{sec:capacity}) and the contribution of each loss component (\Cref{sec:loss-analysis}). A broader discussion of the limitations of our approach is deferred to Appendix~\ref{sec:limitations}.

\subsection{Capacity and Weak-to-Strong Generalization}
\label{sec:capacity}

We explore three generator capacities $G$: \textbf{Adapter} (\method) -- corresponds to our default setup (\Cref{sec:method}); 
\textbf{RNN} -- a lightweight recurrent network which aggregates token-level encodings of $\mathbf{c}$ before the same linear head produces $G(\mathbf{c})$; 
\textbf{Linear} -- an ablation that removes the generator adapter and recurrent aggregator, leaving a single bottleneck linear projection of a pooled context representation (see configuration details in Appendix~\ref{app:hyper}).

In \Cref{fig:capacity}, we compare the performance of generators with varying capacity as the number of demonstrations increases. A simple \textit{linear} generator initially tracks the performance of the ICL teacher, but falls behind as more demonstrations are added, indicating limited ability to capture multi-shot composition. The \textit{RNN} generator improves on ICL and linear baselines for moderate demonstration counts, but eventually stagnates.
In contrast, the \textit{Adapter} generator consistently outperforms alternatives, and its advantage widens with larger demonstration counts. These results highlight the importance of generator capacity for both surpassing the ICL teacher and maintaining compositionality as context scales.

In \Cref{fig:w2s}, we analyze the effects of W2S generalization by grouping MMLU tasks into five difficulty levels ($1$–$5$). 
To obtain these groups, we evaluate the base LLaMA 8B with $16$ shots on each MMLU subtask and sort them by accuracy, partitioning into five bins. 
We then evaluate generators in a $4{\times}4$ setup ($16$ demonstrations in total). Across levels $1$–$3$, all generator variants improve over ICL, with gains ordered by capacity (\textit{Adapter}, \textit{RNN}, and \textit{Linear}).  
At greater difficulty, differences sharpen: the linear generator collapses at level $4$–$5$, \textit{RNN} stagnates and drops at level $5$, 
while \textit{Adapter} continues to improve, achieving the largest gains on the hardest tasks. 
These results show that sufficient generator capacity is crucial for surpassing the ICL teacher and maintaining additive composition as complexity grows. 
\method scales with task difficulty, enhancing W2S generalization on the hardest tasks.

\begin{figure*}[]
  \centering
  \begin{minipage}{0.48\linewidth}
    \centering
    \includegraphics[width=\linewidth]{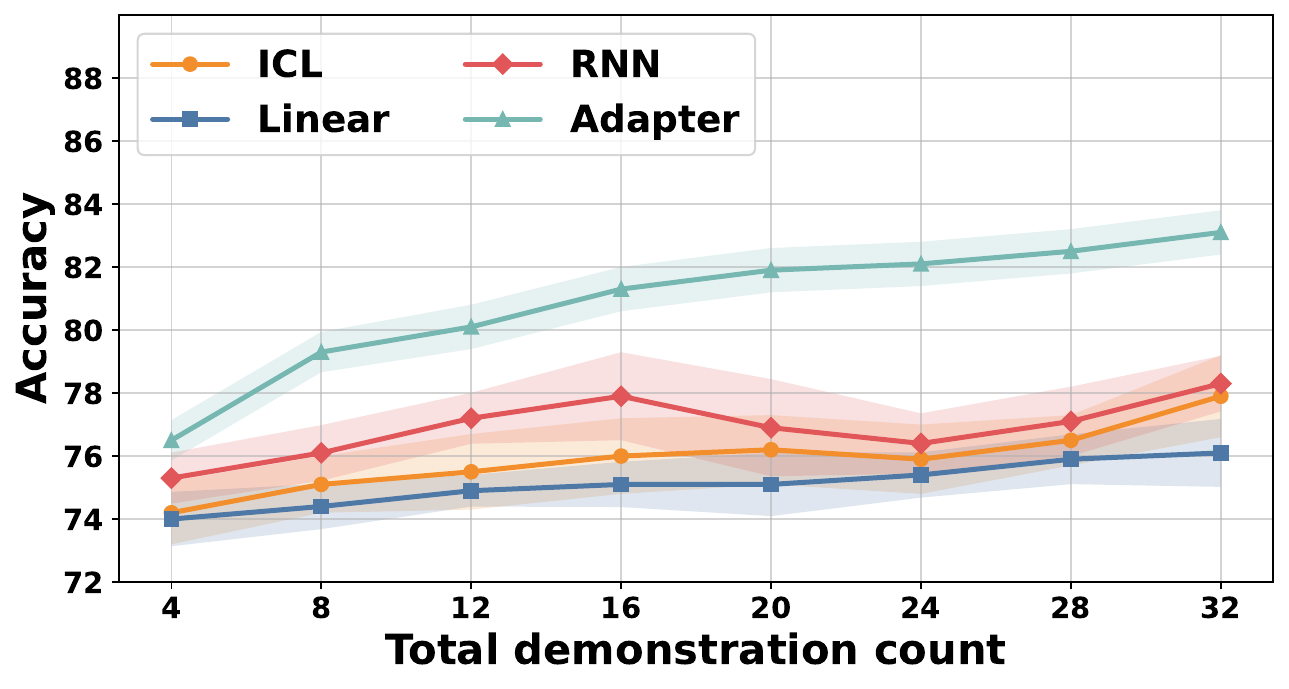}
    \caption{Effect of generator capacity on accuracy with LLaMA 8B on ARC-Challenge. Shaded areas show deviation over $10$ runs.}
    \label{fig:capacity}
  \end{minipage}\hfill
  \begin{minipage}{0.48\linewidth}
    \centering
    \includegraphics[width=\linewidth]{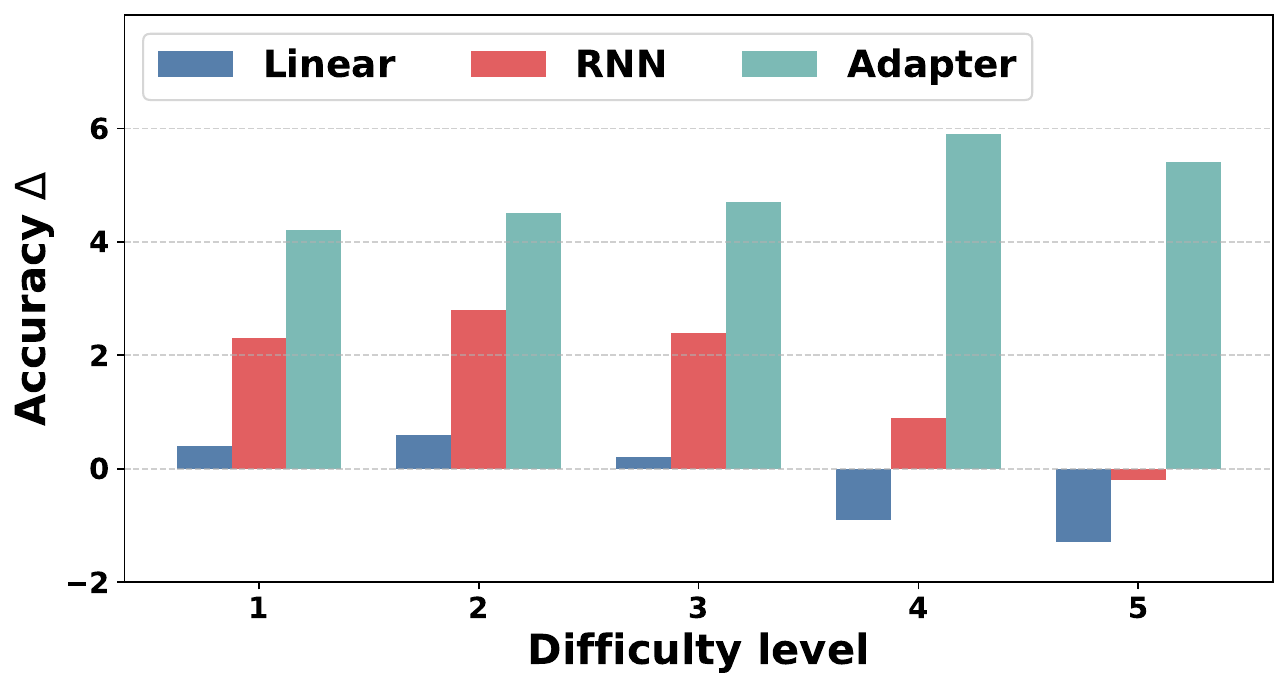}
    \caption{Accuracy deltas of different generators relative to ICL across five MMLU difficulty buckets (stratified by base LLM). Results are averaged over $10$ runs using LLaMA 8B.}
    \label{fig:w2s}
  \end{minipage}
\end{figure*}

\subsection{Importance of Loss Components}
\label{sec:loss-analysis}

To better understand the contribution of each training signal, we perform an ablation study on the loss components of \method. Table \ref{tab:loss-ablation-method-llama8b} reports results on MMLU, ARC-Challenge, and HotpotQA using the LLaMA 8B backbone. We compare the full objective \eqref{eq:objective} to variants with removed components, isolating their effect on overall performance.

We always include $\mathcal{L}_{\text{ST}}$, since weak supervision from the teacher is essential for transferring contextual information into the student. Dropping $\mathcal{L}_{\text{ADD}}$ or $\mathcal{L}_{\text{RECON}}$ degrades performance and shows that they play complementary roles. $\mathcal{L}_{\text{ADD}}$ reduces additivity error ($\eta$) and stabilizes multi-context composition. $\mathcal{L}_{\text{RECON}}$ provides smaller but consistent gains while also enabling context reconstruction. Removing both terms ($\mathcal{L}_{\text{ADD}}$ and $\mathcal{L}_{\text{RECON}}$) leads to the most pronounced drop, confirming that they jointly support the effectiveness of the student. Overall, the full objective yields the strongest results, consistent with the theory in \Cref{sec:theory}.

\begin{table}[t]
\centering
\caption{
Ablation of loss function components.  
Results are on MMLU and ARC ($4{\times}4$ demos) and HotpotQA (EM/F1, pairs).  
We report averages over $10$ seeds.}
\label{tab:loss-ablation-method-llama8b}
\small
\begin{tabular}{lccc}
\toprule
\textbf{Objective} 
& \textbf{MMLU}
& \textbf{ARC} 
& \textbf{HotpotQA (EM/F1)} \\
\midrule
\textbf{Full} 
& $69.1$ & $79.3$ & $71.3/82.3$ \\
\quad$-$\,\(\mathcal{L}_{\text{ADD}}\)
& $64.8$ & $75.7$ & $68.2/79.3$ \\
\quad$-$\,\(\mathcal{L}_{\text{RECON}}\)
& $66.4$ & $77.1$ & $70.2/80.8$ \\
\midrule
\quad$\mathcal{L}_{\text{ST}}$ only
& $61.2$ & $71.8$ & $65.9/76.4$ \\
\bottomrule
\end{tabular}
\end{table}

\section{Related Work}
\label{sec:rw}

\paragraph{ICL and parameterized adaptation.}
Analyses of ICL suggest that models often exploit surface-level cues rather than learning task semantics \citep{min-etal-2022-rethinking}, and that transformers can internally implement simple learning algorithms during inference \citep{xie2022an,akyurek2023what}. In parallel, \emph{parameter-efficient fine-tuning} (PEFT) replaces long prompts with compact trainable modules such as soft/prefix prompts \citep{li-liang-2021-prefix,lester2021prompttuning} or low-rank adapters \citep{hu2022lora}. Recently, \citet{hong2024moicl} proposed mixtures of in-context learners to efficiently combine subsets of demonstrations. Like PEFT, our work keeps inference on the query only, but instead of concatenating demonstrations, we translate them into compositional adapters.

\paragraph{Composing and merging parameters.}
Another line of research explores how to \emph{compose} learned modules or entire models. In the adapter setting, AdapterFusion learns to fuse multiple task adapters non-destructively \citep{pfeiffer2021adapterfusion}, while MAD-X composes language and task adapters for cross-lingual transfer \citep{pfeiffer2020madx}. For full model weights, \emph{model soups} average fine-tuned checkpoints to improve robustness \citep{wortsman2022soups}, and \emph{task vectors} use weight differences to add or subtract task behaviors \citep{ilharco2023editing}. These approaches combine pre-trained modules or models after the fact. In contrast, we generate adapters \emph{on-the-fly} from context and train them explicitly for additivity, such that their composition in parameter space is functionally equivalent to text concatenation.

\paragraph{Generating parameters from context.}
\emph{HyperNetworks} generate parameters of one network using another  \citep{ha2016hypernetworks}. \citet{ansell2021madg} introduced MAD-G, where a hypernetwork generates language adapters conditioned on language embeddings. Subsequent work extended this idea to few-shot adaptation \citep{chen2025generativeadapter}. Our method can be viewed as a hypernetwork explicitly designed for \emph{compositionality}: a generator maps support encodings into LoRA parameters, while compositional distillation and a linearity regularizer enforce that adapter addition in parameter space approximates context concatenation in input space. This aligns with broader efforts to impose linear additive structure on representations, but uniquely emphasizes parameter-space composition as a scalable route to context integration.
\section{Conclusion}
\label{sec:conclusion}

We tackled the challenges of efficiency and instability that arise when adapting LLMs to tasks requiring multiple demonstrations or retrieved passages.
To address these issues, we introduced \method, a meta-learning teacher--student framework that translates contexts into \textit{compositional adapters}.
\method encodes context into adapter parameters that can be algebraically combined, allowing processing of complex queries without input concatenation.
In this way, we enable efficient handling of large context sets by generating adapters independently and composing them in parameter space, reducing inference cost, mitigating long-context degradation, and circumventing context window limitations.
A reconstruction objective promotes safety and security, ensuring that the input context is decodable from the adapter parameters.
\method consistently outperforms ICL and prior generator-based approaches on multi-choice and extractive question answering. Taken together, these results establish compositional context parametrization as a scalable approach for adapting LLMs.

\section*{Impact Statement}
This work proposes encoding contextual information into compositional adapter parameters rather than using long input prompts. Such modular parameterization can reduce inference costs, improve stability in long-context settings, and enable more efficient adaptation of large language models. By structuring context as components in parameter space, the approach may facilitate modular updates and systematic analysis compared to prompt-based adaptation. However, encoding context into parameters introduces privacy and security considerations. If sensitive information is used during training, it may become recoverable from learned adapters. Practical deployment should therefore incorporate safeguards such as controlled access to adapter parameters, data minimization, and privacy-preserving training methods.

\bibliography{literature}

@inproceedings{brown2020language,
 author = {Brown, Tom and Mann, Benjamin and Ryder, Nick and Subbiah, Melanie and Kaplan, Jared D and Dhariwal, Prafulla and Neelakantan, Arvind and Shyam, Pranav and Sastry, Girish and Askell, Amanda and Agarwal, Sandhini and Herbert-Voss, Ariel and Krueger, Gretchen and Henighan, Tom and Child, Rewon and Ramesh, Aditya and Ziegler, Daniel and Wu, Jeffrey and Winter, Clemens and Hesse, Chris and Chen, Mark and Sigler, Eric and Litwin, Mateusz and Gray, Scott and Chess, Benjamin and Clark, Jack and Berner, Christopher and McCandlish, Sam and Radford, Alec and Sutskever, Ilya and Amodei, Dario},
 booktitle = {Advances in Neural Information Processing Systems},
 editor = {H. Larochelle and M. Ranzato and R. Hadsell and M.F. Balcan and H. Lin},
 pages = {1877--1901},
 publisher = {Curran Associates, Inc.},
 title = {Language Models are Few-Shot Learners},
 url = {https://proceedings.neurips.cc/paper_files/paper/2020/file/1457c0d6bfcb4967418bfb8ac142f64a-Paper.pdf},
 volume = {33},
 year = {2020}
}

@article{liu-etal-2024-lost,
    title = "Lost in the Middle: How Language Models Use Long Contexts",
    author = "Liu, Nelson F.  and
      Lin, Kevin  and
      Hewitt, John  and
      Paranjape, Ashwin  and
      Bevilacqua, Michele  and
      Petroni, Fabio  and
      Liang, Percy",
    journal = "Transactions of the Association for Computational Linguistics",
    volume = "12",
    year = "2024",
    address = "Cambridge, MA",
    publisher = "MIT Press",
    url = "https://aclanthology.org/2024.tacl-1.9/",
    doi = "10.1162/tacl_a_00638",
    pages = "157--173"
}

@inproceedings{houlsby2019parameter,
  title = 	 {Parameter-Efficient Transfer Learning for {NLP}},
  author =       {Houlsby, Neil and Giurgiu, Andrei and Jastrzebski, Stanislaw and Morrone, Bruna and De Laroussilhe, Quentin and Gesmundo, Andrea and Attariyan, Mona and Gelly, Sylvain},
  booktitle = 	 {Proceedings of the 36th International Conference on Machine Learning},
  pages = 	 {2790--2799},
  year = 	 {2019},
  editor = 	 {Chaudhuri, Kamalika and Salakhutdinov, Ruslan},
  volume = 	 {97},
  series = 	 {Proceedings of Machine Learning Research},
  month = 	 {09--15 Jun},
  publisher =    {PMLR},
  url = 	 {https://proceedings.mlr.press/v97/houlsby19a.html}
}

@inproceedings{
hu2022lora,
title={Lo{RA}: Low-Rank Adaptation of Large Language Models},
author={Edward J Hu and Yelong Shen and Phillip Wallis and Zeyuan Allen-Zhu and Yuanzhi Li and Shean Wang and Lu Wang and Weizhu Chen},
booktitle={International Conference on Learning Representations},
year={2022},
url={https://openreview.net/forum?id=nZeVKeeFYf9}
}

@inproceedings{
karimi2021compacter,
title={Compacter: Efficient Low-Rank Hypercomplex Adapter Layers},
author={Rabeeh Karimi Mahabadi and James Henderson and Sebastian Ruder},
booktitle={Advances in Neural Information Processing Systems},
editor={A. Beygelzimer and Y. Dauphin and P. Liang and J. Wortman Vaughan},
year={2021},
url={https://openreview.net/forum?id=bqGK5PyI6-N}
}

@inproceedings{
he2022towards,
title={Towards a Unified View of Parameter-Efficient Transfer Learning},
author={Junxian He and Chunting Zhou and Xuezhe Ma and Taylor Berg-Kirkpatrick and Graham Neubig},
booktitle={International Conference on Learning Representations},
year={2022},
url={https://openreview.net/forum?id=0RDcd5Axok}
}

@inproceedings{
bansal2022metaadapters,
title={Meta-Adapters: Parameter Efficient Few-shot Fine-tuning through Meta-Learning},
author={Trapit Bansal and Salaheddin Alzubi and Tong Wang and Jay-Yoon Lee and Andrew McCallum},
booktitle={First Conference on Automated Machine Learning (Main Track)},
year={2022},
url={https://openreview.net/forum?id=BCGNf-prLg5}
}

@inproceedings{chen2025generativeadapter,
  title={Generative Adapter: Contextualizing Language Models in Parameters with A Single Forward Pass},
  author={Chen, Tong and Fang, Hao and Xia, Patrick and Liu, Xiaodong and Van Durme, Benjamin and Zettlemoyer, Luke and Gao, Jianfeng and Cheng, Hao},
  booktitle={International Conference on Learning Representations (ICLR)},
  year={2025},
  url={https://openreview.net/forum?id=bc3sUsS6ck}
}

@inproceedings{
ouyang2022instructgpt,
title={Training language models to follow instructions with human feedback},
author={Long Ouyang and Jeffrey Wu and Xu Jiang and Diogo Almeida and Carroll Wainwright and Pamela Mishkin and Chong Zhang and Sandhini Agarwal and Katarina Slama and Alex Gray and John Schulman and Jacob Hilton and Fraser Kelton and Luke Miller and Maddie Simens and Amanda Askell and Peter Welinder and Paul Christiano and Jan Leike and Ryan Lowe},
booktitle={Advances in Neural Information Processing Systems},
editor={Alice H. Oh and Alekh Agarwal and Danielle Belgrave and Kyunghyun Cho},
year={2022},
url={https://openreview.net/forum?id=TG8KACxEON}
}

@inproceedings{borgeaud2022improving,
  title = 	 {Improving Language Models by Retrieving from Trillions of Tokens},
  author =       {Borgeaud, Sebastian and Mensch, Arthur and Hoffmann, Jordan and Cai, Trevor and Rutherford, Eliza and Millican, Katie and Van Den Driessche, George Bm and Lespiau, Jean-Baptiste and Damoc, Bogdan and Clark, Aidan and De Las Casas, Diego and Guy, Aurelia and Menick, Jacob and Ring, Roman and Hennigan, Tom and Huang, Saffron and Maggiore, Loren and Jones, Chris and Cassirer, Albin and Brock, Andy and Paganini, Michela and Irving, Geoffrey and Vinyals, Oriol and Osindero, Simon and Simonyan, Karen and Rae, Jack and Elsen, Erich and Sifre, Laurent},
  booktitle = 	 {Proceedings of the 39th International Conference on Machine Learning},
  pages = 	 {2206--2240},
  year = 	 {2022},
  editor = 	 {Chaudhuri, Kamalika and Jegelka, Stefanie and Song, Le and Szepesvari, Csaba and Niu, Gang and Sabato, Sivan},
  volume = 	 {162},
  series = 	 {Proceedings of Machine Learning Research},
  month = 	 {17--23 Jul},
  publisher =    {PMLR},
  url = 	 {https://proceedings.mlr.press/v162/borgeaud22a.html},
}

@inproceedings{jacovi2020towards,
    title = "Towards Faithfully Interpretable {NLP} Systems: How Should We Define and Evaluate Faithfulness?",
    author = "Jacovi, Alon  and
      Goldberg, Yoav",
    editor = "Jurafsky, Dan  and
      Chai, Joyce  and
      Schluter, Natalie  and
      Tetreault, Joel",
    booktitle = "Proceedings of the 58th Annual Meeting of the Association for Computational Linguistics",
    month = jul,
    year = "2020",
    address = "Online",
    publisher = "Association for Computational Linguistics",
    url = "https://aclanthology.org/2020.acl-main.386/",
    doi = "10.18653/v1/2020.acl-main.386",
    pages = "4198--4205"
}

@inproceedings{
turpin2023language,
title={Language Models Don't Always Say What They Think: Unfaithful Explanations in Chain-of-Thought Prompting},
author={Miles Turpin and Julian Michael and Ethan Perez and Samuel R. Bowman},
booktitle={Thirty-seventh Conference on Neural Information Processing Systems},
year={2023},
url={https://openreview.net/forum?id=bzs4uPLXvi}
}

@inproceedings{lampinen2022can,
    title = "Can language models learn from explanations in context?",
    author = "Lampinen, Andrew  and
      Dasgupta, Ishita  and
      Chan, Stephanie  and
      Mathewson, Kory  and
      Tessler, Mh  and
      Creswell, Antonia  and
      McClelland, James  and
      Wang, Jane  and
      Hill, Felix",
    editor = "Goldberg, Yoav  and
      Kozareva, Zornitsa  and
      Zhang, Yue",
    booktitle = "Findings of the Association for Computational Linguistics: EMNLP 2022",
    month = dec,
    year = "2022",
    address = "Abu Dhabi, United Arab Emirates",
    publisher = "Association for Computational Linguistics",
    url = "https://aclanthology.org/2022.findings-emnlp.38/",
    doi = "10.18653/v1/2022.findings-emnlp.38",
    pages = "537--563"
}

@inproceedings{devlin-etal-2019-bert,
    title = "{BERT}: Pre-training of Deep Bidirectional Transformers for Language Understanding",
    author = "Devlin, Jacob  and
      Chang, Ming-Wei  and
      Lee, Kenton  and
      Toutanova, Kristina",
    editor = "Burstein, Jill  and
      Doran, Christy  and
      Solorio, Thamar",
    booktitle = "Proceedings of the 2019 Conference of the North {A}merican Chapter of the Association for Computational Linguistics: Human Language Technologies, Volume 1 (Long and Short Papers)",
    month = jun,
    year = "2019",
    address = "Minneapolis, Minnesota",
    publisher = "Association for Computational Linguistics",
    url = "https://aclanthology.org/N19-1423/",
    doi = "10.18653/v1/N19-1423",
    pages = "4171--4186"
}

@article{raffel2020exploring,
  author  = {Colin Raffel and Noam Shazeer and Adam Roberts and Katherine Lee and Sharan Narang and Michael Matena and Yanqi Zhou and Wei Li and Peter J. Liu},
  title   = {Exploring the Limits of Transfer Learning with a Unified Text-to-Text Transformer},
  journal = {Journal of Machine Learning Research},
  year    = {2020},
  volume  = {21},
  number  = {140},
  pages   = {1--67},
  url     = {http://jmlr.org/papers/v21/20-074.html}
}

@inproceedings{min-etal-2022-rethinking,
    title = "Rethinking the Role of Demonstrations: What Makes In-Context Learning Work?",
    author = "Min, Sewon  and
      Lyu, Xinxi  and
      Holtzman, Ari  and
      Artetxe, Mikel  and
      Lewis, Mike  and
      Hajishirzi, Hannaneh  and
      Zettlemoyer, Luke",
    editor = "Goldberg, Yoav  and
      Kozareva, Zornitsa  and
      Zhang, Yue",
    booktitle = "Proceedings of the 2022 Conference on Empirical Methods in Natural Language Processing",
    month = dec,
    year = "2022",
    address = "Abu Dhabi, United Arab Emirates",
    publisher = "Association for Computational Linguistics",
    url = "https://aclanthology.org/2022.emnlp-main.759/",
    doi = "10.18653/v1/2022.emnlp-main.759",
    pages = "11048--11064",
}

@inproceedings{lester2021prompttuning,
    title = "The Power of Scale for Parameter-Efficient Prompt Tuning",
    author = "Lester, Brian  and
      Al-Rfou, Rami  and
      Constant, Noah",
    editor = "Moens, Marie-Francine  and
      Huang, Xuanjing  and
      Specia, Lucia  and
      Yih, Scott Wen-tau",
    booktitle = "Proceedings of the 2021 Conference on Empirical Methods in Natural Language Processing",
    month = nov,
    year = "2021",
    address = "Online and Punta Cana, Dominican Republic",
    publisher = "Association for Computational Linguistics",
    url = "https://aclanthology.org/2021.emnlp-main.243/",
    doi = "10.18653/v1/2021.emnlp-main.243",
    pages = "3045--3059"
}

@inproceedings{pfeiffer2021adapterfusion,
    title = "{A}dapter{F}usion: Non-Destructive Task Composition for Transfer Learning",
    author = {Pfeiffer, Jonas  and
      Kamath, Aishwarya  and
      R{\"u}ckl{\'e}, Andreas  and
      Cho, Kyunghyun  and
      Gurevych, Iryna},
    editor = "Merlo, Paola  and
      Tiedemann, Jorg  and
      Tsarfaty, Reut",
    booktitle = "Proceedings of the 16th Conference of the European Chapter of the Association for Computational Linguistics: Main Volume",
    month = apr,
    year = "2021",
    address = "Online",
    publisher = "Association for Computational Linguistics",
    url = "https://aclanthology.org/2021.eacl-main.39/",
    doi = "10.18653/v1/2021.eacl-main.39",
    pages = "487--503"
}

@inproceedings{pfeiffer2020madx,
    title = "{MAD-X}: {A}n {A}dapter-{B}ased {F}ramework for {M}ulti-{T}ask {C}ross-{L}ingual {T}ransfer",
    author = "Pfeiffer, Jonas  and
      Vuli{\'c}, Ivan  and
      Gurevych, Iryna  and
      Ruder, Sebastian",
    editor = "Webber, Bonnie  and
      Cohn, Trevor  and
      He, Yulan  and
      Liu, Yang",
    booktitle = "Proceedings of the 2020 Conference on Empirical Methods in Natural Language Processing (EMNLP)",
    month = nov,
    year = "2020",
    address = "Online",
    publisher = "Association for Computational Linguistics",
    url = "https://aclanthology.org/2020.emnlp-main.617/",
    doi = "10.18653/v1/2020.emnlp-main.617",
    pages = "7654--7673"
}

@inproceedings{wortsman2022soups,
  title = 	 {Model soups: {A}veraging weights of multiple fine-tuned models improves accuracy without increasing inference time},
  author =       {Wortsman, Mitchell and Ilharco, Gabriel and Gadre, Samir Ya and Roelofs, Rebecca and Gontijo-Lopes, Raphael and Morcos, Ari S and Namkoong, Hongseok and Farhadi, Ali and Carmon, Yair and Kornblith, Simon and Schmidt, Ludwig},
  booktitle = 	 {Proceedings of the 39th International Conference on Machine Learning},
  pages = 	 {23965--23998},
  year = 	 {2022},
  editor = 	 {Chaudhuri, Kamalika and Jegelka, Stefanie and Song, Le and Szepesvari, Csaba and Niu, Gang and Sabato, Sivan},
  volume = 	 {162},
  series = 	 {Proceedings of Machine Learning Research},
  month = 	 {17--23 Jul},
  publisher =    {PMLR},
  url = 	 {https://proceedings.mlr.press/v162/wortsman22a.html}
}

@inproceedings{
ilharco2023editing,
title={Editing models with task arithmetic},
author={Gabriel Ilharco and Marco Tulio Ribeiro and Mitchell Wortsman and Ludwig Schmidt and Hannaneh Hajishirzi and Ali Farhadi},
booktitle={The Eleventh International Conference on Learning Representations },
year={2023},
url={https://openreview.net/forum?id=6t0Kwf8-jrj}
}

@inproceedings{
ha2016hypernetworks,
title={HyperNetworks},
author={David Ha and Andrew M. Dai and Quoc V. Le},
booktitle={International Conference on Learning Representations},
year={2017},
url={https://openreview.net/forum?id=rkpACe1lx}
}

@inproceedings{lewis2020retrieval,
 author = {Lewis, Patrick and Perez, Ethan and Piktus, Aleksandra and Petroni, Fabio and Karpukhin, Vladimir and Goyal, Naman and K\"{u}ttler, Heinrich and Lewis, Mike and Yih, Wen-tau and Rockt\"{a}schel, Tim and Riedel, Sebastian and Kiela, Douwe},
 booktitle = {Advances in Neural Information Processing Systems},
 editor = {H. Larochelle and M. Ranzato and R. Hadsell and M.F. Balcan and H. Lin},
 pages = {9459--9474},
 publisher = {Curran Associates, Inc.},
 title = {Retrieval-Augmented Generation for Knowledge-Intensive NLP Tasks},
 url = {https://proceedings.neurips.cc/paper_files/paper/2020/file/6b493230205f780e1bc26945df7481e5-Paper.pdf},
 volume = {33},
 year = {2020}
}

@inproceedings{hendrycks2021mmlu,
  title        = {Measuring Massive Multitask Language Understanding},
  author       = {Hendrycks, Dan and Burns, Collin and Basart, Steven and Zou, Andy and Mazeika, Mantas and Song, Dawn and Steinhardt, Jacob},
  booktitle    = {ICLR},
  year         = {2021},
  url          = {https://arxiv.org/abs/2009.03300}
}

@article{clark2018arc,
  title        = {Think you have Solved Question Answering? {T}ry {ARC}, the {AI2} Reasoning Challenge},
  author       = {Clark, Peter and Cowhey, Isaac and Etzioni, Oren and Khot, Tushar and Sabharwal, Ashish and Schoenick, Carissa and Tafjord, Oyvind},
  journal      = {arXiv:1803.05457},
  year         = {2018},
  url          = {http://arxiv.org/abs/1803.05457},
}

@inproceedings{yang2018hotpotqa,
    title = "{H}otpot{QA}: A Dataset for Diverse, Explainable Multi-hop Question Answering",
    author = "Yang, Zhilin  and
      Qi, Peng  and
      Zhang, Saizheng  and
      Bengio, Yoshua  and
      Cohen, William  and
      Salakhutdinov, Ruslan  and
      Manning, Christopher D.",
    editor = "Riloff, Ellen  and
      Chiang, David  and
      Hockenmaier, Julia  and
      Tsujii, Jun{'}ichi",
    booktitle = "Proceedings of the 2018 Conference on Empirical Methods in Natural Language Processing",
    month = oct # "-" # nov,
    year = "2018",
    address = "Brussels, Belgium",
    publisher = "Association for Computational Linguistics",
    url = "https://aclanthology.org/D18-1259/",
    doi = "10.18653/v1/D18-1259",
    pages = "2369--2380"
}

@inproceedings{ansell2021madg,
  title        = {MAD-G: Multilingual Adapter Generation for Efficient Cross-Lingual Transfer},
  author       = {Ansell, Alan and Ponti, Edoardo Maria and Pfeiffer, Jonas and Ruder, Sebastian and Glava{\v{s}}, Goran and Vuli{\'c}, Ivan and Korhonen, Anna},
  booktitle    = {Findings of EMNLP},
  year         = {2021},
  url          = {https://aclanthology.org/2021.findings-emnlp.410/}
}

@article{
pfeiffer2023modular,
title={Modular Deep Learning},
author={Jonas Pfeiffer and Sebastian Ruder and Ivan Vuli{\'c} and Edoardo Ponti},
journal={Transactions on Machine Learning Research},
issn={2835-8856},
year={2023},
url={https://openreview.net/forum?id=z9EkXfvxta},
}

@inproceedings{
dherin-etal-2022-neural,
title={Why neural networks find simple solutions:  The many regularizers of geometric complexity},
author={Benoit Dherin and Michael Munn and Mihaela Rosca and David GT Barrett},
booktitle={Advances in Neural Information Processing Systems},
editor={Alice H. Oh and Alekh Agarwal and Danielle Belgrave and Kyunghyun Cho},
year={2022},
url={https://openreview.net/forum?id=-ZPeUAJlkEu}
}

@inproceedings{
lang-etal-2024-theoretical,
title={Theoretical Analysis of Weak-to-Strong Generalization},
author={Hunter Lang and David Sontag and Aravindan Vijayaraghavan},
booktitle={The Thirty-eighth Annual Conference on Neural Information Processing Systems},
year={2024},
url={https://openreview.net/forum?id=HOSh0SKklE}
}

@inproceedings{
jukic-2025-wilda,
title={Disentangling Latent Shifts of In-Context Learning with Weak Supervision},
author={Josip Juki{\'c} and Jan {\v{S}}najder},
booktitle={The Thirty-ninth Annual Conference on Neural Information Processing Systems},
year={2025},
url={https://openreview.net/forum?id=tAq9Gxdhr0}
}

@inproceedings{
xie2022an,
title={An Explanation of In-context Learning as Implicit Bayesian Inference},
author={Sang Michael Xie and Aditi Raghunathan and Percy Liang and Tengyu Ma},
booktitle={International Conference on Learning Representations},
year={2022},
url={https://openreview.net/forum?id=RdJVFCHjUMI}
}

@inproceedings{
akyurek2023what,
title={What learning algorithm is in-context learning? Investigations with linear models},
author={Ekin Aky{\"u}rek and Dale Schuurmans and Jacob Andreas and Tengyu Ma and Denny Zhou},
booktitle={The Eleventh International Conference on Learning Representations },
year={2023},
url={https://openreview.net/forum?id=0g0X4H8yN4I}
}

@inproceedings{hong2024moicl,
    title = "Mixtures of In-Context Learners",
    author = "Hong, Giwon  and
      Van Krieken, Emile  and
      Ponti, Edoardo  and
      Malkin, Nikolay  and
      Minervini, Pasquale",
    editor = "Che, Wanxiang  and
      Nabende, Joyce  and
      Shutova, Ekaterina  and
      Pilehvar, Mohammad Taher",
    booktitle = "Proceedings of the 63rd Annual Meeting of the Association for Computational Linguistics (Volume 1: Long Papers)",
    month = jul,
    year = "2025",
    address = "Vienna, Austria",
    publisher = "Association for Computational Linguistics",
    url = "https://aclanthology.org/2025.acl-long.1277/",
    doi = "10.18653/v1/2025.acl-long.1277",
    pages = "26332--26351",
    ISBN = "979-8-89176-251-0"
}

@misc{llama3,
      title={The Llama 3 Herd of Models}, 
      author={Aaron Grattafiori and Abhimanyu Dubey and Abhinav Jauhri and Abhinav Pandey and Abhishek Kadian and Ahmad Al-Dahle and Aiesha Letman and Akhil Mathur and Alan Schelten and Alex Vaughan and Amy Yang and Angela Fan and Anirudh Goyal and Anthony Hartshorn and Aobo Yang and Archi Mitra and Archie Sravankumar and Artem Korenev and Arthur Hinsvark and Arun Rao and Aston Zhang and Aurelien Rodriguez and Austen Gregerson and Ava Spataru and Baptiste Roziere and Bethany Biron and Binh Tang and Bobbie Chern and Charlotte Caucheteux and Chaya Nayak and Chloe Bi and Chris Marra and Chris McConnell and Christian Keller and Christophe Touret and Chunyang Wu and Corinne Wong and Cristian Canton Ferrer and Cyrus Nikolaidis and Damien Allonsius and Daniel Song and Danielle Pintz and Danny Livshits and Danny Wyatt and David Esiobu and Dhruv Choudhary and Dhruv Mahajan and Diego Garcia-Olano and Diego Perino and Dieuwke Hupkes and Egor Lakomkin and Ehab AlBadawy and Elina Lobanova and Emily Dinan and Eric Michael Smith and Filip Radenovic and Francisco Guzmán and Frank Zhang and Gabriel Synnaeve and Gabrielle Lee and Georgia Lewis Anderson and Govind Thattai and Graeme Nail and Gregoire Mialon and Guan Pang and Guillem Cucurell and Hailey Nguyen and Hannah Korevaar and Hu Xu and Hugo Touvron and Iliyan Zarov and Imanol Arrieta Ibarra and Isabel Kloumann and Ishan Misra and Ivan Evtimov and Jack Zhang and Jade Copet and Jaewon Lee and Jan Geffert and Jana Vranes and Jason Park and Jay Mahadeokar and Jeet Shah and Jelmer van der Linde and Jennifer Billock and Jenny Hong and Jenya Lee and Jeremy Fu and Jianfeng Chi and Jianyu Huang and Jiawen Liu and Jie Wang and Jiecao Yu and Joanna Bitton and Joe Spisak and Jongsoo Park and Joseph Rocca and Joshua Johnstun and Joshua Saxe and Junteng Jia and Kalyan Vasuden Alwala and Karthik Prasad and Kartikeya Upasani and Kate Plawiak and Ke Li and Kenneth Heafield and Kevin Stone and Khalid El-Arini and Krithika Iyer and Kshitiz Malik and Kuenley Chiu and Kunal Bhalla and Kushal Lakhotia and Lauren Rantala-Yeary and Laurens van der Maaten and Lawrence Chen and Liang Tan and Liz Jenkins and Louis Martin and Lovish Madaan and Lubo Malo and Lukas Blecher and Lukas Landzaat and Luke de Oliveira and Madeline Muzzi and Mahesh Pasupuleti and Mannat Singh and Manohar Paluri and Marcin Kardas and Maria Tsimpoukelli and Mathew Oldham and Mathieu Rita and Maya Pavlova and Melanie Kambadur and Mike Lewis and Min Si and Mitesh Kumar Singh and Mona Hassan and Naman Goyal and Narjes Torabi and Nikolay Bashlykov and Nikolay Bogoychev and Niladri Chatterji and Ning Zhang and Olivier Duchenne and Onur Çelebi and Patrick Alrassy and Pengchuan Zhang and Pengwei Li and Petar Vasic and Peter Weng and Prajjwal Bhargava and Pratik Dubal and Praveen Krishnan and Punit Singh Koura and Puxin Xu and Qing He and Qingxiao Dong and Ragavan Srinivasan and Raj Ganapathy and Ramon Calderer and Ricardo Silveira Cabral and Robert Stojnic and Roberta Raileanu and Rohan Maheswari and Rohit Girdhar and Rohit Patel and Romain Sauvestre and Ronnie Polidoro and Roshan Sumbaly and Ross Taylor and Ruan Silva and Rui Hou and Rui Wang and Saghar Hosseini and Sahana Chennabasappa and Sanjay Singh and Sean Bell and Seohyun Sonia Kim and Sergey Edunov and Shaoliang Nie and Sharan Narang and Sharath Raparthy and Sheng Shen and Shengye Wan and Shruti Bhosale and Shun Zhang and Simon Vandenhende and Soumya Batra and Spencer Whitman and Sten Sootla and Stephane Collot and Suchin Gururangan and Sydney Borodinsky and Tamar Herman and Tara Fowler and Tarek Sheasha and Thomas Georgiou and Thomas Scialom and Tobias Speckbacher and Todor Mihaylov and Tong Xiao and Ujjwal Karn and Vedanuj Goswami and Vibhor Gupta and Vignesh Ramanathan and Viktor Kerkez and Vincent Gonguet and Virginie Do and Vish Vogeti and Vítor Albiero and Vladan Petrovic and Weiwei Chu and Wenhan Xiong and Wenyin Fu and Whitney Meers and Xavier Martinet and Xiaodong Wang and Xiaofang Wang and Xiaoqing Ellen Tan and Xide Xia and Xinfeng Xie and Xuchao Jia and Xuewei Wang and Yaelle Goldschlag and Yashesh Gaur and Yasmine Babaei and Yi Wen and Yiwen Song and Yuchen Zhang and Yue Li and Yuning Mao and Zacharie Delpierre Coudert and Zheng Yan and Zhengxing Chen and Zoe Papakipos and Aaditya Singh and Aayushi Srivastava and Abha Jain and Adam Kelsey and Adam Shajnfeld and Adithya Gangidi and Adolfo Victoria and Ahuva Goldstand and Ajay Menon and Ajay Sharma and Alex Boesenberg and Alexei Baevski and Allie Feinstein and Amanda Kallet and Amit Sangani and Amos Teo and Anam Yunus and Andrei Lupu and Andres Alvarado and Andrew Caples and Andrew Gu and Andrew Ho and Andrew Poulton and Andrew Ryan and Ankit Ramchandani and Annie Dong and Annie Franco and Anuj Goyal and Aparajita Saraf and Arkabandhu Chowdhury and Ashley Gabriel and Ashwin Bharambe and Assaf Eisenman and Azadeh Yazdan and Beau James and Ben Maurer and Benjamin Leonhardi and Bernie Huang and Beth Loyd and Beto De Paola and Bhargavi Paranjape and Bing Liu and Bo Wu and Boyu Ni and Braden Hancock and Bram Wasti and Brandon Spence and Brani Stojkovic and Brian Gamido and Britt Montalvo and Carl Parker and Carly Burton and Catalina Mejia and Ce Liu and Changhan Wang and Changkyu Kim and Chao Zhou and Chester Hu and Ching-Hsiang Chu and Chris Cai and Chris Tindal and Christoph Feichtenhofer and Cynthia Gao and Damon Civin and Dana Beaty and Daniel Kreymer and Daniel Li and David Adkins and David Xu and Davide Testuggine and Delia David and Devi Parikh and Diana Liskovich and Didem Foss and Dingkang Wang and Duc Le and Dustin Holland and Edward Dowling and Eissa Jamil and Elaine Montgomery and Eleonora Presani and Emily Hahn and Emily Wood and Eric-Tuan Le and Erik Brinkman and Esteban Arcaute and Evan Dunbar and Evan Smothers and Fei Sun and Felix Kreuk and Feng Tian and Filippos Kokkinos and Firat Ozgenel and Francesco Caggioni and Frank Kanayet and Frank Seide and Gabriela Medina Florez and Gabriella Schwarz and Gada Badeer and Georgia Swee and Gil Halpern and Grant Herman and Grigory Sizov and Guangyi and Zhang and Guna Lakshminarayanan and Hakan Inan and Hamid Shojanazeri and Han Zou and Hannah Wang and Hanwen Zha and Haroun Habeeb and Harrison Rudolph and Helen Suk and Henry Aspegren and Hunter Goldman and Hongyuan Zhan and Ibrahim Damlaj and Igor Molybog and Igor Tufanov and Ilias Leontiadis and Irina-Elena Veliche and Itai Gat and Jake Weissman and James Geboski and James Kohli and Janice Lam and Japhet Asher and Jean-Baptiste Gaya and Jeff Marcus and Jeff Tang and Jennifer Chan and Jenny Zhen and Jeremy Reizenstein and Jeremy Teboul and Jessica Zhong and Jian Jin and Jingyi Yang and Joe Cummings and Jon Carvill and Jon Shepard and Jonathan McPhie and Jonathan Torres and Josh Ginsburg and Junjie Wang and Kai Wu and Kam Hou U and Karan Saxena and Kartikay Khandelwal and Katayoun Zand and Kathy Matosich and Kaushik Veeraraghavan and Kelly Michelena and Keqian Li and Kiran Jagadeesh and Kun Huang and Kunal Chawla and Kyle Huang and Lailin Chen and Lakshya Garg and Lavender A and Leandro Silva and Lee Bell and Lei Zhang and Liangpeng Guo and Licheng Yu and Liron Moshkovich and Luca Wehrstedt and Madian Khabsa and Manav Avalani and Manish Bhatt and Martynas Mankus and Matan Hasson and Matthew Lennie and Matthias Reso and Maxim Groshev and Maxim Naumov and Maya Lathi and Meghan Keneally and Miao Liu and Michael L. Seltzer and Michal Valko and Michelle Restrepo and Mihir Patel and Mik Vyatskov and Mikayel Samvelyan and Mike Clark and Mike Macey and Mike Wang and Miquel Jubert Hermoso and Mo Metanat and Mohammad Rastegari and Munish Bansal and Nandhini Santhanam and Natascha Parks and Natasha White and Navyata Bawa and Nayan Singhal and Nick Egebo and Nicolas Usunier and Nikhil Mehta and Nikolay Pavlovich Laptev and Ning Dong and Norman Cheng and Oleg Chernoguz and Olivia Hart and Omkar Salpekar and Ozlem Kalinli and Parkin Kent and Parth Parekh and Paul Saab and Pavan Balaji and Pedro Rittner and Philip Bontrager and Pierre Roux and Piotr Dollar and Polina Zvyagina and Prashant Ratanchandani and Pritish Yuvraj and Qian Liang and Rachad Alao and Rachel Rodriguez and Rafi Ayub and Raghotham Murthy and Raghu Nayani and Rahul Mitra and Rangaprabhu Parthasarathy and Raymond Li and Rebekkah Hogan and Robin Battey and Rocky Wang and Russ Howes and Ruty Rinott and Sachin Mehta and Sachin Siby and Sai Jayesh Bondu and Samyak Datta and Sara Chugh and Sara Hunt and Sargun Dhillon and Sasha Sidorov and Satadru Pan and Saurabh Mahajan and Saurabh Verma and Seiji Yamamoto and Sharadh Ramaswamy and Shaun Lindsay and Shaun Lindsay and Sheng Feng and Shenghao Lin and Shengxin Cindy Zha and Shishir Patil and Shiva Shankar and Shuqiang Zhang and Shuqiang Zhang and Sinong Wang and Sneha Agarwal and Soji Sajuyigbe and Soumith Chintala and Stephanie Max and Stephen Chen and Steve Kehoe and Steve Satterfield and Sudarshan Govindaprasad and Sumit Gupta and Summer Deng and Sungmin Cho and Sunny Virk and Suraj Subramanian and Sy Choudhury and Sydney Goldman and Tal Remez and Tamar Glaser and Tamara Best and Thilo Koehler and Thomas Robinson and Tianhe Li and Tianjun Zhang and Tim Matthews and Timothy Chou and Tzook Shaked and Varun Vontimitta and Victoria Ajayi and Victoria Montanez and Vijai Mohan and Vinay Satish Kumar and Vishal Mangla and Vlad Ionescu and Vlad Poenaru and Vlad Tiberiu Mihailescu and Vladimir Ivanov and Wei Li and Wenchen Wang and Wenwen Jiang and Wes Bouaziz and Will Constable and Xiaocheng Tang and Xiaojian Wu and Xiaolan Wang and Xilun Wu and Xinbo Gao and Yaniv Kleinman and Yanjun Chen and Ye Hu and Ye Jia and Ye Qi and Yenda Li and Yilin Zhang and Ying Zhang and Yossi Adi and Youngjin Nam and Yu and Wang and Yu Zhao and Yuchen Hao and Yundi Qian and Yunlu Li and Yuzi He and Zach Rait and Zachary DeVito and Zef Rosnbrick and Zhaoduo Wen and Zhenyu Yang and Zhiwei Zhao and Zhiyu Ma},
      year={2024},
      eprint={2407.21783},
      archivePrefix={arXiv},
      primaryClass={cs.AI},
      url={https://arxiv.org/abs/2407.21783}, 
}

@misc{qwen2025,
      title={Qwen2.5 Technical Report}, 
      author={An Yang and Baosong Yang and Beichen Zhang and Binyuan Hui and Bo Zheng and Bowen Yu and Chengyuan Li and Dayiheng Liu and Fei Huang and Haoran Wei and Huan Lin and Jian Yang and Jianhong Tu and Jianwei Zhang and Jianxin Yang and Jiaxi Yang and Jingren Zhou and Junyang Lin and Kai Dang and Keming Lu and Keqin Bao and Kexin Yang and Le Yu and Mei Li and Mingfeng Xue and Pei Zhang and Qin Zhu and Rui Men and Runji Lin and Tianhao Li and Tianyi Tang and Tingyu Xia and Xingzhang Ren and Xuancheng Ren and Yang Fan and Yang Su and Yichang Zhang and Yu Wan and Yuqiong Liu and Zeyu Cui and Zhenru Zhang and Zihan Qiu},
      year={2025},
      eprint={2412.15115},
      archivePrefix={arXiv},
      primaryClass={cs.CL},
      url={https://arxiv.org/abs/2412.15115}, 
}

@inproceedings{cho2014gru,
  title     = {Learning Phrase Representations using {RNN} Encoder--Decoder for Statistical Machine Translation},
  author    = {Cho, Kyunghyun and van Merri{\"e}nboer, Bart and Gulcehre, Caglar and Bahdanau, Dzmitry and Bougares, Fethi and Schwenk, Holger and Bengio, Yoshua},
  booktitle = {Proceedings of the 2014 Conference on Empirical Methods in Natural Language Processing (EMNLP)},
  pages     = {1724--1734},
  year      = {2014},
  publisher = {Association for Computational Linguistics},
  doi       = {10.3115/v1/D14-1179},
  url       = {https://aclanthology.org/D14-1179}
}

@inproceedings{li-liang-2021-prefix,
    title = "Prefix-Tuning: Optimizing Continuous Prompts for Generation",
    author = "Li, Xiang Lisa  and
      Liang, Percy",
    editor = "Zong, Chengqing  and
      Xia, Fei  and
      Li, Wenjie  and
      Navigli, Roberto",
    booktitle = "Proceedings of the 59th Annual Meeting of the Association for Computational Linguistics and the 11th International Joint Conference on Natural Language Processing (Volume 1: Long Papers)",
    month = aug,
    year = "2021",
    address = "Online",
    publisher = "Association for Computational Linguistics",
    url = "https://aclanthology.org/2021.acl-long.353/",
    doi = "10.18653/v1/2021.acl-long.353",
    pages = "4582--4597",
}

@inproceedings{schick-schutze-2021-exploiting,
    title = "Exploiting Cloze-Questions for Few-Shot Text Classification and Natural Language Inference",
    author = {Schick, Timo  and
      Sch{\"u}tze, Hinrich},
    editor = "Merlo, Paola  and
      Tiedemann, Jorg  and
      Tsarfaty, Reut",
    booktitle = "Proceedings of the 16th Conference of the European Chapter of the Association for Computational Linguistics: Main Volume",
    month = apr,
    year = "2021",
    address = "Online",
    publisher = "Association for Computational Linguistics",
    url = "https://aclanthology.org/2021.eacl-main.20/",
    doi = "10.18653/v1/2021.eacl-main.20",
    pages = "255--269",
}

@inproceedings{reimers-gurevych-2019-sentence,
    title = "Sentence-{BERT}: Sentence Embeddings using {S}iamese {BERT}-Networks",
    author = "Reimers, Nils  and
      Gurevych, Iryna",
    editor = "Inui, Kentaro  and
      Jiang, Jing  and
      Ng, Vincent  and
      Wan, Xiaojun",
    booktitle = "Proceedings of the 2019 Conference on Empirical Methods in Natural Language Processing and the 9th International Joint Conference on Natural Language Processing (EMNLP-IJCNLP)",
    month = nov,
    year = "2019",
    address = "Hong Kong, China",
    publisher = "Association for Computational Linguistics",
    url = "https://aclanthology.org/D19-1410/",
    doi = "10.18653/v1/D19-1410",
    pages = "3982--3992"
}
\bibliographystyle{icml2026}

\newpage
\appendix
\onecolumn
\appendix

\section{Compositionality Bound}
\label{app:proof}

\subsection{Proof of Theorem~\ref{thm:compositionality}}

Let $\boldsymbol{\phi}_1 = G(\mathbf{c}_1)$, 
$\boldsymbol{\phi}_2 = G(\mathbf{c}_2)$, 
and $\boldsymbol{\phi}_{12} = G([\mathbf{c}_1;\mathbf{c}_2])$.  
We aim to bound
\[
\big\|\,\mathbf{f}_S^{\boldsymbol{\phi}_1+\boldsymbol{\phi}_2}(\mathbf{q})
-\mathbf{f}_T([\mathbf{c}_1;\mathbf{c}_2;\mathbf{q}])\,\big\|_2.
\]

The Euclidean space $(\mathbb{R}^{|V|},\|\cdot\|_2)$ satisfies the triangle inequality: for any $a,b,c\in\mathbb{R}^{|V|}$,
\begin{equation}\label{eq:triangle}
\|a-c\|_2 \;\le\; \|a-b\|_2 \;+\; \|b-c\|_2,
\end{equation}
where we also use absolute homogeneity (symmetry) of the norm, i.e., 
$\|b-c\|_2=\|c-b\|_2$ since $\|{-}v\|_2=\|v\|_2$.
Apply \eqref{eq:triangle} with
\[
a=\mathbf{f}_S^{\boldsymbol{\phi}_1+\boldsymbol{\phi}_2}(\mathbf{q}), \quad
b=\mathbf{f}_S^{\boldsymbol{\phi}_{12}}(\mathbf{q}), \quad
c=\mathbf{f}_T([\mathbf{c}_1;\mathbf{c}_2;\mathbf{q}]),
\]
to obtain
\begin{align}
&\|\mathbf{f}_S^{\boldsymbol{\phi}_1+\boldsymbol{\phi}_2}(\mathbf{q})
-\mathbf{f}_T([\mathbf{c}_1;\mathbf{c}_2;\mathbf{q}])\|_2 \nonumber\\
&\qquad \le 
\underbrace{\|\mathbf{f}_S^{\boldsymbol{\phi}_1+\boldsymbol{\phi}_2}(\mathbf{q})
-\mathbf{f}_S^{\boldsymbol{\phi}_{12}}(\mathbf{q})\|_2}_{\text{Term (I)}} \;+\;
\underbrace{\|\mathbf{f}_S^{\boldsymbol{\phi}_{12}}(\mathbf{q})
-\mathbf{f}_T([\mathbf{c}_1;\mathbf{c}_2;\mathbf{q}])\|_2}_{\text{Term (II)}}.
\label{eq:decompose}
\end{align}

By parameter sensitivity \eqref{eq:parameter_sensitivity} and generator additivity error \eqref{eq:generator_additivity_error},
\begin{equation}\label{eq:termI}
\text{Term (I)} \;\le\; L \,\|\boldsymbol{\phi}_1+\boldsymbol{\phi}_2-\boldsymbol{\phi}_{12}\|_2
\;=\; L\,\eta.
\end{equation}

Apply the student--teacher error \eqref{eq:student_teacher_error} to the concatenated context $[\mathbf{c}_1;\mathbf{c}_2]$:
\begin{equation}\label{eq:termII}
\text{Term (II)} \;\le\; \epsilon([\mathbf{c}_1;\mathbf{c}_2]).
\end{equation}

Combining \eqref{eq:decompose}, \eqref{eq:termI}, and \eqref{eq:termII} yields
\[
\|\mathbf{f}_S^{\boldsymbol{\phi}_1+\boldsymbol{\phi}_2}(\mathbf{q})
-\mathbf{f}_T([\mathbf{c}_1;\mathbf{c}_2;\mathbf{q}])\|_2
\;\le\; L \eta + \epsilon([\mathbf{c}_1;\mathbf{c}_2]),
\]
which proves Theorem~\ref{thm:compositionality}. \qed

\begin{corollary}[Extension to $k$ contexts]
\label{cor:k-context}
By induction, Theorem~\ref{thm:compositionality} extends to any sequence of 
contexts $\mathbf{c}_1,\ldots,\mathbf{c}_k$ with corresponding adapters 
$\boldsymbol{\phi}_i = G(\mathbf{c}_i)$:
\[
\|\mathbf{f}_S^{\sum_{i=1}^k \boldsymbol{\phi}_i}(\mathbf{q})
- \mathbf{f}_T([\mathbf{c}_1;\ldots;\mathbf{c}_k;\mathbf{q}])\|_2
\;\le\; (k-1)L\eta \;+\; \epsilon([\mathbf{c}_1;\ldots;\mathbf{c}_k]).
\]
\end{corollary}

\subsection{Geometric interpretation}
\label{app:geometry}

Adapters $\boldsymbol{\phi}=G(\mathbf{c})$ can be viewed as vectors in the parameter space $\Phi$.  
In the ideal case, $G$ embeds contexts into a flat subspace where concatenation in input space corresponds exactly to vector addition in parameter space.  
The deviation terms then acquire a geometric meaning:  
$\eta$ measures the generator’s departure from additivity,  
and $\epsilon([\mathbf{c}_1;\mathbf{c}_2])$ captures the student’s misalignment with the teacher on concatenated contexts.  
The Lipschitz constant $L$ governs how deviations in adapter space propagate into the model’s output space.  
Compositionality can therefore be understood as flattening the contextual geometry into a nearly linear embedding, with small $\eta$ and $\epsilon$ ensuring that adapter addition faithfully mirrors context concatenation.

\section{Experimental Setup and Hyperparameters}
\label{app:hyper}

\subsection{Data}
For MMLU \citep{hendrycks2021mmlu}, which comprises $59$ subject-specific subsets, we report accuracy averaged across all subsets. We train a \emph{single generator} jointly across all subsets. This ensures that the generator is shared across domains rather than tuned separately per subject. For weak supervision during generator optimization, we sample queries and contexts from the validation split without using ground-truth labels, while demonstrations are drawn (with labels) from the training split. Evaluation is performed on the held-out test split.

For ARC-Challenge \citep{clark2018arc}, we follow the same setup: unlabeled validation examples provide queries and contexts for weak supervision, labeled demonstrations are drawn from the training set, and performance is reported on the test set.

For HotpotQA \citep{yang2018hotpotqa}, we restrict context inputs to the relevant gold supporting passages associated with each question. We sample $10$k examples stratified across difficulty levels, and partition them into $5$k for training, $3$k for validation, and $2$k for testing.

\subsection{Models}
\label{app:models}
We summarize the main characteristics of the base LMs used in our experiments in \Cref{tab:models}, together with the sizes of the student and generator adapters. We use the \texttt{bfloat16} half-precision format for all model parameters.

\subsection{Methods}
\label{app:methods}
For GenAda \citep{chen2025generativeadapter} and \wilda \citep{jukic-2025-wilda}, we use the default hyperparameters and follow the training procedures described in the respective papers. To ensure comparability, we adopt an identical configuration for adapter sizes and target modules across all methods; these design choices are detailed in the following section.

\begin{table}[t]
\centering
\caption{Parameter counts for base models and our framework. 
Student parameters correspond to LoRA adapters attached to the base LM, 
while generator parameters denote the generator network itself. 
Generator adapter counts indicate the size of parameters produced per context. 
Total trainable parameters are the sum of student and generator parameters.}
\label{tab:models}
\begin{tabular}{lcccc}
\toprule
\textbf{Model} & \textbf{Base LLM} & \textbf{Student adapter} & \textbf{Generator adapter} & \textbf{Generator (total)} \\
\midrule
Llama~3.1 (8B) & $8.03$B & $21$M & $42$M & $210$M  \\
Llama~3.2 (3B) & $3.19$B & $4.2$M & $8.4$M & $42$M \\
Qwen~2.5 (7.6B) & $7.61$B & $20$M & $40$M & $180$M \\
\bottomrule
\end{tabular}
\end{table}

\subsection{Hyperparameters}

\paragraph{Optimization.}  
For each dataset and model combination, we train the generator parameters with AdamW (weight decay $0.01$) for $10$ epochs.
The learning rate is set to $10^{-4}$ with $5$\% linear warmup followed by cosine decay.

\paragraph{Loss weights.}
We set $\lambda_{\text{ST}}=1.0$, $\lambda_{\text{ADD}}=0.5$, and $\lambda_{\text{RECON}}=0.1$.
These values balance (i) fidelity to teacher logits through student–teacher alignment, (ii) enforcement of compositionality via additivity regularization, and (iii) auxiliary reconstruction for stability and faithfulness, while avoiding over-regularization.

\paragraph{Adapter configuration.}
We insert LoRA adapters into all attention projections ($q$, $k$, $v$, and $o$) as well as into the MLP down- and up-projection layers. For the generator, we use a uniform rank of $32$ across all modules, while for the student, we use rank of $16$. Following standard practice, each LoRA module applies a scaling factor $\alpha$, such that the effective weight update is $\frac{\alpha}{r}AB$. We set $\alpha=32$ for generator adapters and $\alpha=16$ for student adapters, ensuring balanced contribution of low-rank updates relative to their respective ranks.

\paragraph{Reconstruction under composition.}
In the compositional setting, where multiple adapters are generated for different contexts and summed, we compute the reconstruction F1 score in a permutation-invariant fashion. This adjustment is necessary because the generator is explicitly incentivized to be commutative, i.e., $G([c_1;c_2]) \approx G(c_1)+G(c_2) = G(c_2)+G(c_1)$. Concretely, for multiple contexts, we evaluate reconstruction across all possible permutations and report the score corresponding to the most successful ordering. This ensures that the metric reflects content preservation rather than sensitivity to input order.

\paragraph{Generator variants.}
In addition to the default \textit{Adapter} generator used in \method, we consider two alternatives for ablation. The \textit{RNN} variant employs a lightweight two-layer gated recurrent unit (GRU; \citealp{cho2014gru}) with hidden size $256$ to aggregate token-level context representations before projecting them into a compact latent space, followed by an up-projection to the full set of student adapter parameters. The \textit{Linear} variant removes both the GRU and the generator adapter; instead, the pooled context representation is passed through a single linear bottleneck and then expanded directly into the student adapter parameter space.

\subsection{Computing Infrastructure}
We conducted our experiments on a mix of local and cluster resources. Local training was performed on a workstation with an AMD Ryzen Threadripper 3970X 32-Core CPU, 256GB RAM, and NVIDIA GeForce RTX 3090 GPUs. Larger-scale runs were executed on a compute cluster equipped with NVIDIA A100 GPUs.

\section{Additional Experiments}
\label{app:experiments}

\subsection{Long-Form Summarization}
Beyond question answering, we also evaluate whether the representations learned by \method generalize to long-form generation tasks. Its reconstruction objective already requires producing full demonstrations and multi-paragraph passages (\Cref{tab:compas_reconstruction}), exercising long-range generation in training.

To test generalization capabilities beyond QA, we additionally evaluate \method on XSUM, a standard abstractive summarization benchmark.
Results are reported in Table~\ref{tab:xsum}.
\method achieves the strongest performance across all ROUGE metrics, substantially outperforming ICL and other baselines. This indicates that the learned context-to-parameter mapping transfers to tasks that require extended, structured generation, supporting the use of \method as a general parametrization mechanism.

\begin{table}
\centering
\caption{XSUM summarization with LLaMA-3.1 8B.
We report ROUGE-1 (R1), ROUGE-2 (R2), and ROUGE-L (RL) F1. R1 and R2 measure unigram and bigram overlap with the reference summary, while RL measures the longest common subsequence. Higher values indicate closer alignment with the reference and better summary quality.}
\label{tab:xsum}
\begin{tabular}{lccc}
\toprule
\textbf{Method} & \textbf{R1} $\uparrow$ & \textbf{R2} $\uparrow$ & \textbf{RL} $\uparrow$ \\
\midrule
ICL         & 36.1 & 15.6 & 28.4 \\
PBFT        & 38.3 & 18.1 & 31.6 \\
GenAda      & 33.4 & 13.1 & 25.6 \\
\method     & \textbf{41.8} & \textbf{22.8} & \textbf{33.5} \\
\bottomrule
\end{tabular}
\end{table}

\subsection{Supervised and Prompt-Based Baselines}

We further evaluate supervised and parametric baselines:
(i) pattern-based fine-tuning (PBFT), optimized with a cloze-style language modeling objective \citep{schick-schutze-2021-exploiting};
(ii) a linear SFT head trained on top of the frozen backbone;
(iii) prefix-tuning with the same demonstrations \citep{li-liang-2021-prefix}; and
(iv) an ICL variant that retrieves the top-$k$ most similar demonstrations for each query using SBERT embeddings \citep{reimers-gurevych-2019-sentence}.

Results for MMLU and ARC are reported in \Cref{tab:supervised_baselines}.
Across both benchmarks, \method consistently outperforms all supervised and prompt-based baselines, supporting that its improvements arise from the context-to-parameter mechanism rather than differences in data, supervision, or training procedure.

\begin{table}[t]
\centering
\caption{Supervised and prompt-based baselines with LLaMA-3.1 8B.
We report mean accuracy on MMLU and ARC over $10$ runs with $16$ demonstrations in total.
\method is evaluated with $4\times4$ composition. Standard deviations are shown in subscript, and the best results are in bold.}
\label{tab:supervised_baselines}
\begin{tabular}{lcc}
\toprule
\textbf{Method} & \textbf{MMLU} & \textbf{ARC} \\
\midrule
ICL (16-shot)              & $64.2_{1.8}$ & $74.2_{1.6}$ \\
ICL + prefix-tuning        & $66.1_{1.4}$ & $74.0_{1.3}$ \\
ICL + similarity selection & $68.1_{0.6}$ & $75.5_{0.2}$ \\
PBFT (labels only)         & $66.8_{1.1}$ & $75.6_{1.0}$ \\
Linear SFT head            & $62.1_{1.5}$ & $71.3_{1.4}$ \\
\method (4$\times$4)       & $\mathbf{72.2_{0.3}}$ & $\mathbf{81.3_{0.3}}$ \\
\bottomrule
\end{tabular}
\end{table}

\subsection{Positional Encoding and Order Sensitivity}
\label{sec:position}

By design, our generator enforces \emph{commutativity}, i.e., 
$G([\mathbf{c}_1;\mathbf{c}_2]) \approx G(\mathbf{c}_1) + G(\mathbf{c}_2) = G(\mathbf{c}_2) + G(\mathbf{c}_1)$, 
which is desirable for tasks where order invariance is beneficial and directly reduces $\eta$. 
The trade-off is that strict commutativity removes the ability to encode order. 

As a \emph{proof of concept}, we conduct an experiment on \textsc{HotpotQA} using pairs of supporting contexts. 
We augment the generator with \emph{slot embeddings}, i.e., for slot $i$ we add a learnable vector $\mathbf{s}_i$ to the corresponding adapter delta:  
\begin{equation}
\tilde{\boldsymbol{\phi}}_i = G(\mathbf{c}_i) + \mathbf{s}_i,
\qquad
\boldsymbol{\phi} = \sum_i \tilde{\boldsymbol{\phi}}_i.
\label{eq:slot-embedding}
\end{equation}  
This introduces positional variance while preserving linear composition, thereby allowing the model to distinguish 
$[\mathbf{c}_1;\mathbf{c}_2]$ from $[\mathbf{c}_2;\mathbf{c}_1]$. 

Results in \Cref{tab:positional_ablation} suggest that order sensitivity can be layered on top of the commutative backbone when tasks require it, though we leave a systematic study to future work.

\begin{table}[]
\centering
\caption{Position-aware reconstruction evaluation.
We report positional ROUGE-L F1 for context pair permutations from HotpotQA for the commutative baseline without positional encoding (No-PE) and a simple position-aware variant (PE-Tag), and slot-based positional embeddings (Slot-PE).
Scores are mean with standard deviation as subscripts over $10$ runs.}
\label{tab:positional_ablation}
\begin{tabular}{lccc}
\toprule
 & \textbf{LLaMA-3.1 8B} & \textbf{LLaMA-3.2 3B} & \textbf{Qwen-2.5 7B} \\
\midrule
No-PE (commutative linear) & $62.5_{2.0}$ & $50.1_{2.2}$ & $60.4_{2.1}$ \\
Slot-PE      & $87.6_{0.9}$ & $71.1_{2.6}$ & $85.0_{1.4}$ \\
\bottomrule
\end{tabular}
\end{table}

\subsection{Compositional Scaling}
\label{sec:comp_scaling}

A central design choice in \method is how context is partitioned into blocks prior to adapter generation.
Block size controls both how much information is encoded into a single adapter and how many such adapters must be composed at inference time, and therefore directly influences both expressiveness and compositional stability.
We next examine how this tradeoff affects accuracy and additivity as the total number of demonstrations increases.

\Cref{tab:block_granularity} shows how block structure affects both task accuracy and compositional fidelity. For a fixed number of demonstrations, \method reaches its best performance with intermediate block sizes (approximately 4--8 demonstrations per block), which yield both higher MMLU accuracy and lower additivity error $\eta$.

When blocks become too large (e.g., $2\times16$ or $1\times16$), the generator must encode many demonstrations jointly, which weakens the linear structure of the latent parameterization and increases additivity error.
Conversely, when blocks become too small (e.g., $16\times1$), each block provides a limited contextual signal, leading to slightly reduced accuracy and moderately higher $\eta$.

As the total number of demonstrations increases, accuracy steadily improves, demonstrating that \method benefits from additional evidence.
In practice, the maximum feasible block size is limited by GPU memory, since larger blocks require the generator to encode more tokens jointly.
Within these limits, using moderately sized blocks provides a favorable tradeoff between representation stability and information density.

Finally, long-context ICL baselines can only be evaluated reliably up to $32$ demonstrations due to memory constraints.
In contrast, \method decouples inference cost from context length, enabling evaluation with more demonstrations on the same hardware through compositional adapter merging.

\begin{table}
\centering
\caption{Effect of block granularity and scaling on MMLU accuracy and additivity error $\eta$ with LLaMA 8B.
We vary the block structure for a fixed total number of demonstrations and report accuracy and generator additivity error.}
\label{tab:block_granularity}
\begin{tabular}{llcc}
\toprule
\textbf{Total demos} & \textbf{Method} & \textbf{Accuracy} & $\boldsymbol{\eta}$ $\downarrow$ \\
\midrule
16 & ICL (16-shot)        & 64.2 & -- \\
16 & \method (4$\times$4)  & 72.2 & 0.08 \\
16 & \method (2$\times$8)  & 70.9 & 0.11 \\
16 & \method (1$\times$16) & 69.4 & 0.15 \\
16 & \method (16$\times$1) & 71.3 & 0.09 \\
\midrule
32 & ICL (32-shot)     & 65.5 & -- \\
32 & \method (8$\times$4)  & 73.1 & 0.07 \\
32 & \method (4$\times$8)  & 74.0 & 0.09 \\
32 & \method (2$\times$16) & 72.4 & 0.19 \\
\midrule
40 & \method (5$\times$8)  & 74.3 & 0.10 \\
48 & \method (6$\times$8)  & 74.9 & 0.09 \\
56 & \method (7$\times$8)  & 75.4 & 0.12 \\
64 & \method (8$\times$8)  & 76.2 & 0.16 \\
\bottomrule
\end{tabular}
\end{table}

\subsection{Robustness to Noisy Context}

In practical retrieval-based systems, the contexts provided to a model are often imperfect: demonstrations may contain incorrect labels, multiple sources may disagree about the same example, or retrieved passages may be unrelated to the query. We therefore evaluate how \method behaves under controlled perturbations that explicitly simulate these failure modes and compare it directly to standard ICL under identical corruption patterns.

We introduce three types of degradation at inference time. \emph{Noisy demonstrations} are generated by randomly flipping the labels of a fixed fraction of the in-context examples ($12.5$\% and $25$\% of the $16$ demonstrations), while keeping inputs unchanged. This simulates annotation errors or unreliable retrieval from weakly supervised corpora. \emph{Contradictory demonstrations} are constructed by duplicating a subset of examples and assigning them conflicting labels: for $4$ of the $16$ demonstrations, the same input appears twice across contexts, once with the correct label and once with an incorrect label. This explicitly creates inconsistent supervision inside the context and mimics disagreement between retrieved sources or conflicting knowledge bases. \emph{Off-topic contamination} is introduced by appending unrelated paragraphs sampled from HotpotQA to the end of the context, increasing the number of irrelevant tokens without changing the gold demonstrations. This simulates retrieval failures in which non-informative or spurious documents are mixed into the prompt. All results are reported as averages over 10 independent runs with different random corruption patterns.

Table~\ref{tab:noise_mmlu} reports accuracy on MMLU as the demonstration noise increases. Both methods degrade as noise grows, but \method consistently retains a clear performance margin. Notably, the gap widens under stronger label corruption and remains substantial under contradictory supervision. This suggests that compositional context encoding is more robust to internally inconsistent evidence than raw concatenation: while ICL treats all tokens uniformly, \method compresses each demonstration into a latent representation, reducing the impact of spurious or conflicting signals.

Table~\ref{tab:noise_hotpot} reports results on HotpotQA as irrelevant passages are injected into the context. ICL performance degrades steadily as more off-topic content is introduced, indicating sensitivity to prompt dilution and distraction effects. In contrast, \method remains comparatively stable and preserves most of its clean-context accuracy even as irrelevant text grows. This behavior is consistent with the hypothesis that encoding context into parameters rather than raw tokens makes inference less sensitive to unrelated input.

Taken together, these experiments demonstrate that composition-based context parametrization improves robustness to realistic retrieval noise. Even when demonstrations contain errors, contradictions, or irrelevant information, \method degrades more gracefully than ICL, supporting additive composition as a stabilizing mechanism under imperfect and noisy context conditions.

\begin{table}
\centering
\caption{Robustness under noisy and contradictory demonstrations on MMLU with LLaMA 8B (mean over $10$ runs).}
\label{tab:noise_mmlu}
\begin{tabular}{lcccc}
\toprule
\textbf{Method} 
& \textbf{Clean} 
& \textbf{12.5\% noise} 
& \textbf{25\% noise} 
& \textbf{Contradictory} \\
\midrule
ICL        & 66.5 & 62.1 & 58.7 & 57.9 \\
\method    & \textbf{72.2} & \textbf{71.3} & \textbf{69.4} & \textbf{68.4} \\
\bottomrule
\end{tabular}
\end{table}

\begin{table}
\centering
\caption{HotpotQA under off-topic context contamination with LLaMA 8B.
Accuracy with irrelevant paragraphs appended to the context (mean over $10$ runs).}
\label{tab:noise_hotpot}
\begin{tabular}{lccc}
\toprule
\textbf{Method} 
& \textbf{Clean} 
& \textbf{+1 off-topic} 
& \textbf{+2 off-topic} \\
\midrule
ICL      & 82.0 & 79.4 & 77.6 \\
\method  & 82.9 & 82.3 & 81.7 \\
\bottomrule
\end{tabular}
\end{table}

\subsection{Efficiency}
\label{sec:efficiency}

We compare the computational and memory efficiency of \method to standard ICL prompting using LLaMA 8B on ARC-Challenge. \Cref{tab:efficiency} reports FLOPs speedup (relative to ICL with the same number of demonstrations $k$) and peak inference memory. Results are averaged over $10$ runs on a sample of $1$k test queries. As expected, the memory cost of ICL grows linearly with the number of demonstrations, since the model must re-encode all tokens at inference. In contrast, \method amortizes context encoding into a one-shot parameter generation step, so inference depends only on the query length. This yields substantial speedups that increase with $k$: $2.2\times$ at $4$-shot, $3.7\times$ at $12$-shot, and $4.1\times$ at $16$-shot. Memory usage shows a similar trend, with ICL increasing steadily as more demonstrations are added, while \method remains nearly constant across different $k$. This demonstrates that \method not only improves accuracy, but also yields more efficient and scalable inference by decoupling compute and memory from context length.

Additionally, Table~\ref{tab:compute_overview} compares absolute inference and training compute across methods.
The methods differ primarily in their training regimes.
\wilda is inexpensive for a single demonstration configuration, but does not amortize: each new context requires a separate fine-tuning run. 
GenAda incurs a large one-time training cost to learn a generator.
\method lies between these extremes, training a generator once at moderate cost and reusing it across contexts.
In cumulative terms, \wilda{}’s total cost grows linearly with the number of demonstration configurations and overtakes \method after approximately $17$ contexts, beyond which repeated fine-tuning becomes more expensive than the one-time generator training of \method.
At the same time, \method is approximately $13\times$ cheaper to train than GenAda, while achieving comparable inference-time efficiency.
Thus, \method provides a favorable tradeoff between repeated fine-tuning and large upfront training cost, combining amortization with moderate training overhead.

\begin{table}
\centering
\caption{Efficiency comparison of ICL vs.\ \method with LLaMA 8B. 
We report speedups in FLOPs for \method inference relative to ICL with $k$ demonstrations, and peak inference memory (GB). 
Results are averaged over $10$ runs on ARC-Challenge. 
\method replaces long-context prompts with adapters, reducing both compute and memory usage.}
\label{tab:efficiency}
\begin{adjustbox}{max width=\textwidth}
\begin{tabular}{lcccccccc}
\toprule
& \multicolumn{4}{c}{\textbf{Speedup (FLOPs rel.\ to ICL@$k$)} $\uparrow$} & \multicolumn{4}{c}{\textbf{Peak memory (GB)} $\downarrow$} \\
\cmidrule(lr){2-5}\cmidrule(lr){6-9}
\textbf{Method} & \textbf{4} & \textbf{8} & \textbf{12} & \textbf{16} & \textbf{4} & \textbf{8} & \textbf{12} & \textbf{16} \\
\midrule
ICL (prompting)   & $1.0\times$ & $1.0\times$ & $1.0\times$ & $1.0\times$ & 22.1 & 27.3 & 33.5 & 37.9 \\
\method           & $2.2\times$ & $3.1\times$ & $3.7\times$ & $4.1\times$ & 17.2 & 17.9 & 18.1 & 19.4 \\
\bottomrule
\end{tabular}
\end{adjustbox}
\end{table}

\begin{table}
\centering
\caption{Absolute compute comparison with LLaMA 8B.
We report absolute inference-time compute per query and training compute in TFLOPs on the ARC-Challenge dataset using $16$ demonstrations in total with $4\times4$ composition for \method.
\method trains a new adapter for each demonstration configuration, whereas GenAda and \method train a generator once and amortize across different contexts.}
\label{tab:compute_overview}
\begin{tabular}{lccl}
\toprule
\textbf{Method} 
& \textbf{Inference TFLOPs / query} 
& \textbf{Training TFLOPs} 
& \textbf{Regime} \\
\midrule
ICL (prompting) 
& $4.0$ 
& --
& No training \\

\wilda 
& $0.7$ 
& $1.4\times10^{5}$ 
& Per demo set \\

GenAda 
& $0.9$ 
& $3.1\times10^{7}$
& One-time \\

\method 
& $0.8$ 
& $2.4\times10^{6}$ 
& One-time \\
\bottomrule
\end{tabular}
\end{table}

\section{Prompt Templates}
\label{app:prompts}

\subsection{Templates for multi-choice question answering}

\begin{center}
\begin{tcolorbox}[colback=gray!5!white, colframe=gray!75!black, width=\textwidth]
\begin{center}
    \textbf{Generic prompt template MMLU and ARC-Challenge} \\[4pt]
\end{center}
\textbf{Demonstrations:}
\begin{verbatim}
Question: {Previous Question 1}
Answer choices:
 (A: {Choice A1}),
 (B: {Choice B1}),
 (C: {Choice C1}),
 (D: {Choice D1})
Answer: (Correct Answer 1)

Question: {Previous Question 2}
Answer choices:
(A: {Choice A2}),
(B: {Choice B2}),
(C: {Choice C2}),
(D: {Choice D2})
Answer: (Correct Answer 2)
...
\end{verbatim}
\textbf{Query:}
\begin{verbatim}
Question: {Current Question}
Answer choices:
(A: {Choice A}),
(B: {Choice B}),
(C: {Choice C}),
(D: {Choice D})
Answer: (
\end{verbatim}
\end{tcolorbox}
\end{center}

\subsection{MMLU Examples}

\begin{tcolorbox}[colback=gray!5!white, colframe=gray!75!black, width=\textwidth, breakable]
\begin{center}
    \textbf{Example for MMLU \texttt{abstract\_algebra}} \\[4pt]
\end{center}

\textbf{Demonstrations:}

\begin{lstlisting}[basicstyle=\ttfamily\small, breaklines=true]
Question: Find the maximum possible order for an element of S_n for n = 10.
Answer choices: 
(A: 6), 
(B: 12), 
(C: 30), 
(D: 105)
Answer: (C: 30)

Question: Compute the product in the given ring. (2,3)(3,5) in Z_5 x Z_9
Answer choices:
(A: (1,1)), 
(B: (3,1)), 
(C: (1,6)), 
(D: (3,6))
Answer: (D: (3,6))
\end{lstlisting}

\textbf{Query:}

\begin{lstlisting}[basicstyle=\ttfamily\small, breaklines=true]
Question: If (G, .) is a group such that (ab)^-1 = a^-1b^-1 
for all a, b in G, then G is a/an
Answer choices:
(A: commutative semigroup), 
(B: abelian group), 
(C: non-abelian group), 
(D: None of these)
Answer: (
\end{lstlisting}

\end{tcolorbox}

\subsection{ARC-Challenge Examples}
\noindent
\begin{tcolorbox}[breakable, colback=gray!5!white, colframe=gray!75!black, width=\textwidth]
\begin{center}
\textbf{Example for ARC-Challenge}\par
\end{center}
\medskip

\textbf{Demonstrations:}

\begin{lstlisting}[basicstyle=\ttfamily\small, breaklines=true]
Question: Based on their locations in the periodic table, 
which element has chemical properties most similar 
to those of calcium, Ca?
Answer choices: 
(A: beryllium, Be),
(B: potassium, K),
(C: titanium, Ti),
(D: yttrium, Y)
Answer: (A: beryllium, Be)

Question: Which term best describes the life cycle of an insect 
that reaches the adult stage without being a pupa?
Answer choices:
(A: incomplete metamorphosis),
(B: complete metamorphosis),
(C: alternation of generations),
(D: spontaneous mutation)
Answer: (A: incomplete metamorphosis)
\end{lstlisting}

\textbf{Query:}

\begin{lstlisting}[basicstyle=\ttfamily\small, breaklines=true]
Question: Which property of a mineral can be determined 
just by looking at it?
Answer choices:
(A: luster),
(B: mass),
(C: weight),
(D: hardness)
Answer: (
\end{lstlisting}
\end{tcolorbox}

\subsection{HotpotQA Example}

\begin{tcolorbox}[breakable, colback=gray!5!white, colframe=gray!75!black, width=\textwidth]
\begin{center}
\textbf{Example for HotpotQA} \\[4pt]
\end{center}

\textbf{Question:} \\
Who invented the type of script used in autographs? \\[4pt]

\textbf{Supporting Context:}

\begin{lstlisting}[basicstyle=\ttfamily\small, breaklines=true]
(Cuneiform script):
Cuneiform script, one of the earliest systems of writing, 
was invented by the Sumerians. It is distinguished by its 
wedge-shaped marks on clay tablets, made by means of a blunt 
reed for a stylus. The name "cuneiform" itself simply means 
"wedge shaped".

(Autograph in Assyriology):
An autograph in Assyriology is the hand-copy of a cuneiform 
clay-tablet. Producing an autograph is often the first step of 
a tablet's archaeological interpretation and the autograph is 
frequently the authoritative form that is published as source 
material. Autographing the text is followed by transliteration, 
transcription and translation.
\end{lstlisting}

\textbf{Answer:} Sumerians

\end{tcolorbox}

\section{Limitations}
\label{sec:limitations}

\paragraph{Order-insensitivity of composition.}
In \method, parameter-space composition is strictly commutative and associative, since adapters are summed without regard to order. This differs from textual concatenation, where $(c_A, c_B)$ and $(c_B, c_A)$ may yield different interpretations. While our results show benefits of order-robustness (mitigating prompt-order sensitivity), the lack of positional information may reduce expressivity in tasks where sequence order is essential, such as instruction chaining or narrative reasoning. In additional experiments, we provide a proof of concept showing that positional information can be encoded within the generator, demonstrating feasibility but leaving systematic exploration to future work.

\paragraph{Generator robustness.}
Our generator is trained on a meta-distribution of contexts and shows cross-domain transfer, but generalization to entirely novel domains or reasoning styles is not guaranteed. In particular, low-resource or highly specialized tasks may expose brittle adaptation.

\paragraph{Linearity vs. expressivity.}
The framework enforces strict linear compositionality of adapters to ensure interpretability and additivity guarantees. However, some tasks inherently require non-linear interactions between contexts (e.g., resolving contradictions or multi-hop reasoning). In such cases, \method may face a trade-off between maintaining modularity and achieving task-optimal integration.

\paragraph{Scalability and efficiency trade-offs.}
Although \method replaces long contexts with compact adapters, the generator itself introduces a large up-projection layer whose parameter count grows with the size of the student adapters. This overhead is modest relative to the base LM but still significant. Scaling to even larger backbones or more complex adapter schemes may therefore incur efficiency and memory costs that partially offset the gains from shorter inference contexts.

\section{Reproducibility Statement}

Along with the description of our method in \Cref{sec:method} and the experimental setup in \Cref{app:hyper}, we provide comprehensive details to ensure reproducibility. All datasets used (MMLU, ARC-Challenge, HotpotQA) are publicly available and described in \Cref{sec:setup}. Model configurations, optimization settings, and training schedules are fully specified in \Cref{app:hyper}, along with the specified computing infrastructure. Theoretical results, including the proof of Theorem~\ref{thm:compositionality} and its corollary, are presented in \Cref{app:proof}. Prompt templates for all benchmarks are provided in \Cref{app:prompts}. In addition, we include code in the supplementary material and will publish the repository publicly after the review process.



\end{document}